\documentclass[sigconf,10pt]{acmart}

\usepackage{subfigure}                  
\usepackage{multirow}                   
\usepackage{enumitem}                   
\usepackage{balance}                    
\usepackage{graphicx}
\usepackage{caption}
\usepackage{adjustbox}
\usepackage{algorithmicx}
\usepackage{url}
\usepackage[utf8]{inputenc}
\usepackage{comment}
\usepackage{minted}
\usepackage{xcolor}
\usepackage{tcolorbox}
\usepackage{soul}
\usepackage{color, colortbl}
\definecolor{codebg}{rgb}{0.95, 0.95, 0.92}

\usepackage{amsmath}
\usepackage[linesnumbered,ruled,vlined]{algorithm2e}

\usepackage{array}
\usepackage{tabularray}
\usepackage{soul}

\usepackage{tikz}
\newcommand*\circled[1]{\tikz[baseline=(char.base)]{
            \node[shape=circle,draw,inner sep=0.8pt] (char) {#1};}}

\newcommand{\kc}[1]{{\textcolor{brown}{#1}}}

\newcommand{\jk}[1]{{\textcolor{red}{\textbf{JK: #1}}}}

\newcommand{\rev}[1]{{\textcolor{red}{{#1}}}}
\newcommand{\system}{\textit{Tazza}}
\newcommand{\name}{\textit{Tazza}}

\usepackage{xr}
\begin{document}

\setcopyright{none}
\settopmatter{printacmref=false} 
\renewcommand\footnotetextcopyrightpermission[1]{} 
\pagestyle{plain}

\makeatletter

\title{Neural Network Shuffling for Secure and Private Federated Learning}

\author{Kichang Lee, Jaeho Jin, JaeYeon Park$^*$, Songkuk Kim, and JeongGil Ko}
\affiliation{College of Computing, Yonsei University, $^*$ Department of Mobile Systems Engineering, Dankook University\country{}}

\begin{abstract}
%
%
Federated learning enables decentralized model training without sharing raw data from individual user/client devices, thereby preserving data privacy. However, federated learning systems still face significant security challenges, from data leakage via gradient inversion and model poisoning from malicious clients. Existing solutions often address these issues in isolation, which compromises either system robustness or model accuracy. In this paper, we present \name{}, a secure and efficient federated learning framework designed to tackle both challenges simultaneously. By leveraging the permutation equivariance and invariance properties inherent in modern neural network architectures, \name{} enhances resilience against diverse poisoning attacks, including noise injection and backdoor attacks, while ensuring data confidentiality and maintaining high model accuracy. Evaluations using various datasets and real-world embedded platforms show that \name{} achieves robust defense with up to 6.40$\times$ improved computational efficiency compared to alternative schemes, without compromising the model performance.



\end{abstract}
\maketitle
\section{Introduction}
\label{sec:intro}

Federated learning is a transformative paradigm that trains machine learning models in decentralized environments to preserve the privacy of distributed mobile and embedded data while enhancing personalized experiences~\cite{mcmahan2017communication, shin2024effective}. In such systems, clients locally train their models and share only model updates, rather than raw data, with a central server, which aggregates model updates from many distributed clients to build a robust global model.

However, despite its data privacy-preserving design, federated learning still faces notable security challenges. Specifically, its reliance on distributed, heterogeneous client data makes it susceptible to malicious model updates that compromise system integrity~\cite{fang2020local,elhattab2023robust}. Additionally, honest-but-curious servers can analyze shared model updates to infer sensitive data, thereby breaching confidentiality~\cite{elhattab2024pastel,feng2020pmf}. These issues are especially critical in ubiquitous computing scenarios, such as daily activity tracking and personal health monitoring, where sensitive data is abundant and the risk is heightened by the installation of third-party apps~\cite{lee2025mind,lee2023exploiting}. In summary, federated learning must contend with two primary attack vectors: \textit{integrity attacks} from malicious clients and \textit{confidentiality attacks} from curious servers.

Integrity attacks arise when malicious clients disrupt federated learning by sharing tampered updates. These attacks include injecting noise~\cite{fang2020local}, using incorrect training labels~\cite{biggio2012poisoning,fung2020limitations}, or embedding adversarial patterns~\cite{bagdasaryan2020backdoor}. Such disruptions significantly impede global model convergence and performance. To counter these threats, researchers have developed defenses like filtering out anomalous updates using statistical priors~\cite{yin2018byzantine} and adopting robust aggregation strategies~\cite{blanchard2017machine,cao2020fltrust}. On the other hand, confidentiality attacks occur when a curious server tries to infer sensitive client data from the shared model updates. Common countermeasures include applying differential privacy techniques~\cite{abadi2016deep,dwork2014algorithmic} or using gradient pruning~\cite{zhu2019deep}. However, these methods can compromise the convergence of the global model due to the alteration of the updates. This paper aims to provide a comprehensive solution to these intertwined challenges in federated learning.

We emphasize that \textit{ensuring both \textbf{integrity} and \textbf{confidentiality}} is critical in federated learning, particularly for mobile and IoT scenarios. With large-scale participation from distributed and easy-to-physically-alter clients, it is unrealistic to assume that all participants act in good faith. Thus, robust mechanisms are needed to defend against malicious clients that compromise the integrity of the system as a whole. Equally important is safeguarding sensitive data collected from mobile/IoT devices, such as health information~\cite{park2020heartquake,ouyang2024admarker}, daily activities~\cite{Park21sugo,fu2024magspy,yun2024powdew}, or environmental metrics~\cite{xie2023mozart,sun2020alexa} to maintain trust and meet regulations.


However, simultaneously addressing integrity and confidentiality poses unique challenges. Traditional confidentiality measures (e.g., differential privacy) obscure model updates to protect privacy, but hinder the detection of malicious patterns such as backdoor attacks or label poisoning. Conversely, integrity defenses focus on identifying and mitigating these threats yet may conflict with confidentiality measures, complicating the design of systems that meet both objectives. This complexity is exacerbated by the non-i.i.d. nature of real-world mobile and IoT data, where update diversity makes it harder to distinguish legitimate and malicious updates, especially when confidentiality defenses show inaccuracies. These trade-offs pose significant barriers to ensuring robust security and privacy without compromising global model performance~\cite{kabir2024flshield}. Despite the critical importance of these issues, few works simultaneously address both threats while preserving performance. Tackling this intertwined challenge is imperative for real-world federated learning applications, where system reliability and user trust depend on balanced, effective defenses.

In this paper, we address the challenges of ensuring both integrity and confidentiality in federated learning by leveraging the permutation equivariant and invariant properties of neural networks (c.f., Sec~\ref{subsec:pinn}). Permutation equivariance refers to the property where the output of a function changes consistently with the order of its input, reflecting identical permutation. Whereas, permutation invariance implies that a function's output remains unchanged regardless of the order of its input. These properties are shared in various neural network architectures, such as multilayer perceptrons, recurrent neural networks, and Transformers.

Leveraging these insights, we propose \name{}, a novel framework that ensures both confidentiality and integrity in federated learning through a novel weight shuffling module and a shuffled model validation process. \name{} features a weight shuffling module that rearranges neural network parameters based on a predefined rule, intentionally obscuring the original model while preserving permutation equivariant and invariant properties. This approach effectively protects sensitive information while remaining computationally efficient for mobile and embedded devices. Additionally, the shuffled model validation process confirms that the rearranged model’s outputs remain consistent and accurate, thereby maintaining its overall performance. We further leverage similarity metrics from this validation to cluster models and isolate malicious updates from benign ones.

Nevertheless, shuffling model parameters without compromising desired properties requires a deep understanding of neural network operations. While many architectures exhibit permutation equivariant and invariant properties, their underlying mechanisms heavily vary. Thus, we introduce tailored algorithms for representative model architectures, demonstrating how permutation-based techniques can be effectively harnessed for practical federated learning.

Overall, we present \name{} as a secure, privacy-preserving, and accurate federated learning framework that addresses both integrity and confidentiality challenges, fostering trust and enabling widespread ubiquitous computing for mobile and IoT services. Backed by a solid mathematical foundation and validated through extensive evaluations, \name{} maintains high accuracy while mitigating attacks, achieving up to a 6.4$\times$ computational speedup compared to alternative defense schemes. Our key contributions are as follows:

\begin{itemize}[leftmargin=*]
    \item Building on mathematical evidence, we propose an effective and efficient model-securing mechanism for federated learning. This mechanism leverages the permutation equivariance and invariance properties of neural networks through \textit{Weight Shuffling} and \textit{Shuffled Model Validation}.

    \item By seamlessly integrating Weight Shuffling and Shuffled Model Validation, we present \name{}, a privacy-preserving and secure federated learning framework specifically designed for mobile and embedded scenarios, achieving robust security without compromising model performance.

    \item We conduct an extensive evaluation of \name{} across diverse datasets, model architectures, and attack scenarios. Our results demonstrate that \name{} effectively isolates malicious model updates from benign ones to counter integrity attacks, while its model parameter shuffling mechanism robustly mitigates confidentiality attacks.

\end{itemize}
\section{Background and Related Works}
\label{sec:background}

\subsection{Security Threats in Federated Learning}
\label{subsec:stfl}

We begin by discussing background information on integrity and confidentiality attacks in federated learning, along with foundational details on the concepts of permutation equivariance and permutation invariance.

\vspace{1ex}

\noindent\textbf{Integrity Attacks.} Malicious clients can compromise federated learning systems by tampering with local model weights. For example, in data poisoning attacks, adversaries alter data-label pairs (e.g., label flipping), while model poisoning involves noise injection or parameter scaling to distort model outcomes~\cite{fang2020local}. Additionally, backdoor attacks train models to output undesired results when triggered by specific patterns~\cite{bagdasaryan2020backdoor}. We collectively term these attacks as \textit{integrity attacks} in this work.

To mitigate such attacks, prior work has proposed secure aggregation techniques that leverage statistical priors or similarity-based approaches. Statistical methods such as Median~\cite{xie2018generalized} and Trimmed-Mean~\cite{yin2018byzantine} treat malicious updates as outliers, while similarity-based methods such as Krum~\cite{blanchard2017machine} and FLTrust~\cite{cao2020fltrust} select updates based on Euclidean or cosine similarity to a trusted model. However, these defenses can discard valuable information, potentially leading to suboptimal global model performance.

\vspace{1ex}
\noindent\textbf{Confidentiality Attacks.} Another significant security threat in federated learning systems is confidentiality attacks, which aim to infer and reveal local training data at the federated learning server. Zhu et al. demonstrated that local data can be reconstructed by comparing the initial model with client-updated parameters using L-BFGS optimization~\cite{zhu2019deep}, and, Geiping et al. introduced the Inverting Gradient attack, which refines reconstruction by incorporating cosine similarity and total variation measures~\cite{geiping2020inverting}.

\begin{figure*}[t!]
    \centering
    \includegraphics[width=0.9\linewidth]{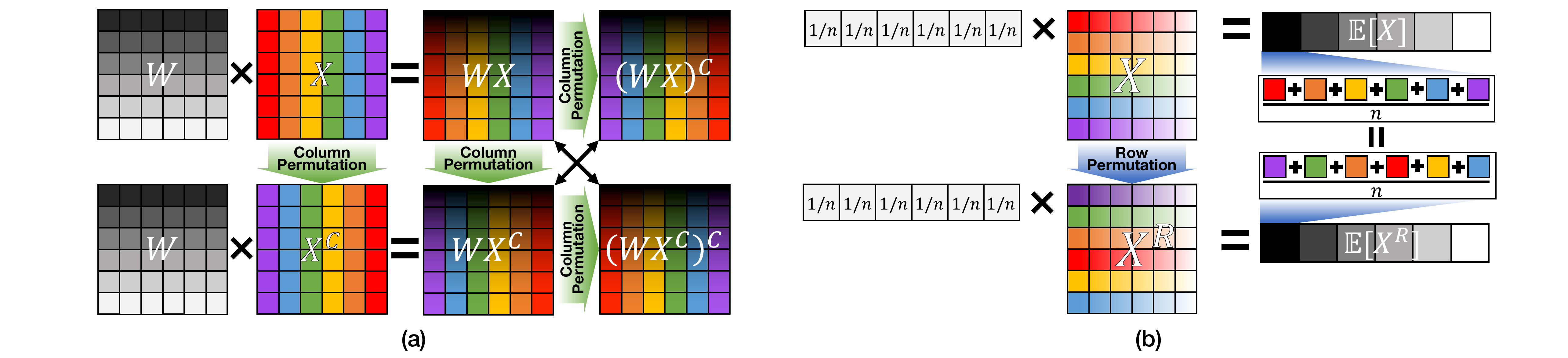}
    \vspace{-4ex}
    \caption{(a) Visualization of the permutation equivariance property of matrix multiplication. The output's column order follows the order of the input columns. (b) Visualization of the permutation invariance property of the average function. Results are not affected by the altered order of input rows. Best viewed in color.}
    \vspace{-2ex}
    \label{fig:pepi}
\end{figure*}

To counter confidentiality attacks, widely adopted approaches include applying differential privacy~\cite{abadi2016deep}, which adds controlled noise to model updates, and gradient pruning~\cite{zhu2019deep}, which suppresses sensitive information in gradients during local training. However, these defense mechanisms often degrade model performance, as they manipulate original updates in ways that hinder convergence and accuracy~\cite{huang2021evaluating}. Furthermore, such approaches often necessitate repetitive computations during local training, which not only delay the federated training operations but also impose additional computational burden on local devices. 

\subsection{Permutation Equivariance/Invariance}
\label{subsec:pinn}

We now explore the concepts of permutation equivariance and permutation invariance, which are fundamental to the framework presented in this work.

\vspace{1ex}
\noindent \textbf{Permutation Equivariance} is the property where a function’s output order mirrors the input order. Figure~\ref{fig:pepi}~(a) illustrates this using matrix multiplication with \(C\) as a column-reversed matrix. When the columns of input $X$ are reversed (\(X \rightarrow X^C\)), the outputs \(WX\) and \(WX^C\) differ only in column order, matching that of the input (i.e., $WX^{C}=(WX)^{C}$).



Formally, if a function \(f(\cdot)\) satisfies permutation equivariance, for a given input \(X = \{x_1, x_2, \dots, x_N\}\) and a permutation function \(\pi(i): \{1, 2, \ldots, N\} \to \{1, 2, \ldots, N\}\), where \(\pi(i)\) is a bijective mapping, the lemma $\pi(f(X)) = f(\pi(X))$ holds. Applying this to matrix multiplication, function \(f(X) = WX\) satisfies permutation equivariance with respect to the column order. Consequently, \(WX^C\) (i.e., shuffling order of input) is identical to \((WX)^C\) (i.e., shuffling order of output).

\vspace{1ex}


\noindent \textbf{Permutation Invariance} describes a function that produces identical results regardless of the order of its input. For example, a function that computes the average of the elements in a given array is permutation invariant because the arrangement of elements does not influence the outcome. Figure~\ref{fig:pepi}~(b) illustrates this concept using the average operation in the context of matrix multiplication, where \(n\) represents the number of rows and \(R\) denotes a row-permuted version of the input. As shown, despite reordering the rows, the final result remains unchanged (i.e., \(\mathbb{E}[X] = \mathbb{E}[X^R]\)). Without loss of generality, if a function \(f(\cdot)\) satisfies permutation invariance, then for a given input \(X\) and a permutation function \(\pi(i)\), the lemma $f(X) = f(\pi(X))$ holds.

Note that matrix multiplication, in general, is \textit{not} permutation invariant. Specifically, if weights used in multiplication are not uniform (e.g., \(\frac{1}{n}\) for all elements, as in the averaging operation in Figure~\ref{fig:pepi} (b)), the result will differ depending on row order. This is because such weights can be interpreted as a weighted sum, being sensitive to the input arrangement, emphasizing the importance of an exquisite shuffling operation in a neural network to retain the property.

\vspace{1ex}
\noindent \textbf{Neural Networks and Shuffling.} Modern neural networks exhibit a \textit{semi}-permutation equivariant and invariant property, often producing nearly identical training results even when the input features (e.g., image pixels) or labels are shuffled~\cite{tolstikhin2021mlp, naseer2021intriguing}. This behavior stems from the dominance of matrix operations in neural network computation~\cite{zhang2019compiler, kim2022fast}, which maintain output semantics under reordering.


\begin{figure}[t!]
    \centering
    \includegraphics[width=\linewidth]{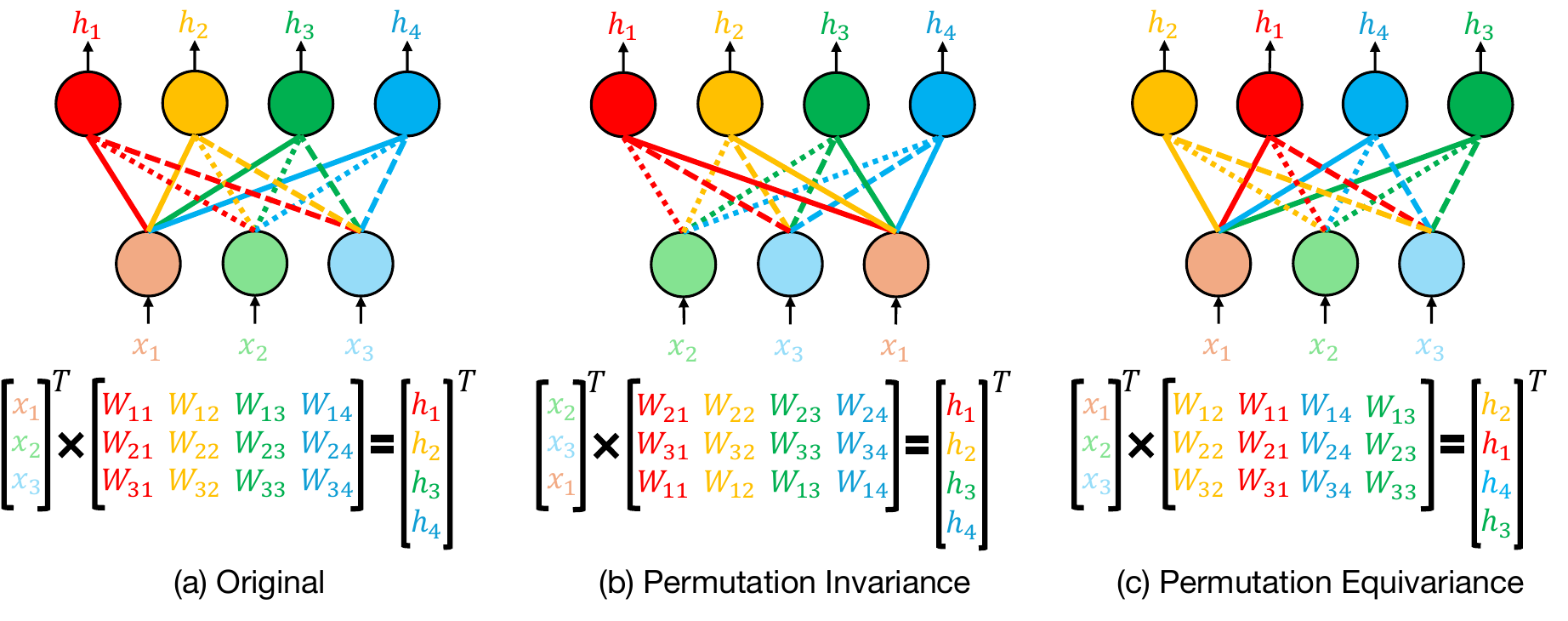}
    \vspace{-5ex}
    \caption{Visual examples of permutation invariance and equivariance properties. (a) Original network architecture. (b) Row-permuted operations, where input  ({$x_{1}$, $x_{2}$, $x_{3}$}) is rearranged to ({$x_{2}$, $x_{3}$, $x_{1}$}). (c) Column-permuted operations, where output is reordered.}
    \label{fig:perm_vis}
\end{figure}


Figure~\ref{fig:perm_vis} illustrates this property, presenting the (a) the base architecture, (b) row-permutation of inputs (preserving outputs—permutation invariance), and (c) column-permutation of weights (reordering outputs—permutation equivariance). These properties enable model computations to remain valid even with shuffled data or weights.

Our \textit{weight-shuffling} approach builds on this insight but differs fundamentally from shuffle-based differential privacy (DP) schemes~\cite{girgis2021shuffled}. Shuffle-model DP anonymizes \textit{who} sent a message by randomly reordering locally randomized messages \textit{across} clients~\cite{cheu2021differential, murakami2025augmented, kon2024netshuffle}. In contrast, we deterministically permute weights \textit{within} each client to conceal \textit{what} the parameters encode, without additional infrastructure.

While prior works explore parameter permutation~\cite{mozaffari2022fedperm}, they often assume a trusted external shuffler and focus solely on obfuscation. By contrast, we provide a rigorous, layer-wise analysis showing that deterministic client-side weight shuffling preserves both linear and nonlinear operations. This transforms weight permutation into a dual-purpose primitive that ensures both privacy and integrity. Through shuffled validation and cluster-aware aggregation, \name{} uses this mechanism not only for obfuscation but as a foundational defense against malicious behaviors—achieving security without compromising accuracy or efficiency.

\section{Feasibility Study and Core Idea}
\label{sec:feasibility}

We conduct a preliminary study showing that data permutation during neural network training can preserve both data privacy and model performance. Building on these observations, we introduce the core idea behind \name{}.

\begin{figure}
    \centering
    \vspace{-3ex}
    \includegraphics[width=.9\linewidth]{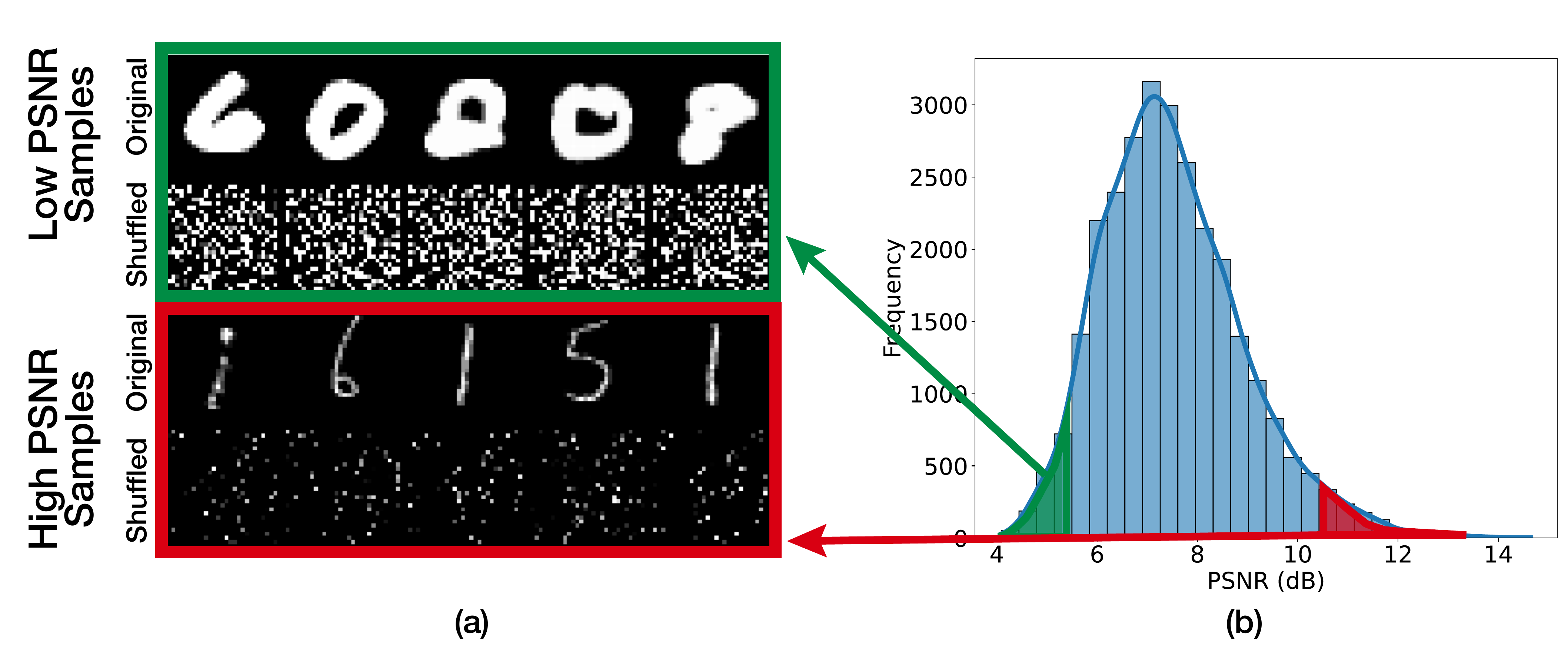}
    \vspace{-3ex}
    \caption{(a) Bottom-5 (top green box) and Top-5 (bottom red box) PSNR samples from the MNIST dataset. (b) PSNR distribution between the original and shuffled images from the MNIST dataset (using 60K images).}
    \vspace{-2ex}
    \label{fig:feasibility-a}
\end{figure}

\vspace{1ex}
\noindent \textbf{Feasibility Study.}
To examine whether shuffling can enhance data privacy, we measured the Peak Signal-to-Noise Ratio (PSNR) between original and shuffled MNIST samples, where each image is permuted using a fixed pixel order. Figure~\ref{fig:feasibility-a} visualizes examples and the PSNR distribution for all 60{,}000 samples. Most shuffled images yield low PSNR values (6–8 dB), and even the top-5 highest-PSNR examples remain semantically unintelligible, confirming that shuffling effectively obscures sensitive content.

\begin{figure}
    \centering
    \includegraphics[width=0.9\linewidth]{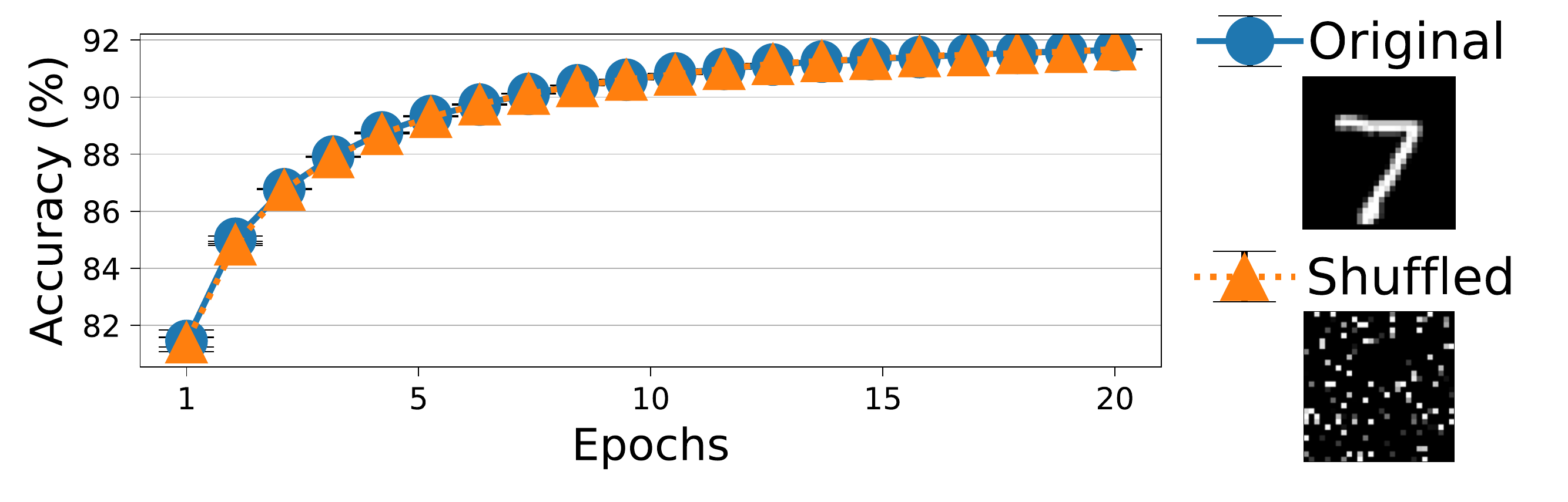}
    \vspace{-4ex}
    \caption{Average accuracy of models trained with original data and shuffled data for 10 random seeds.}
    \vspace{-3ex}
    \label{fig:feasibility-b}
\end{figure}

We then trained a 3-layer MLP for 20 epochs on both original and shuffled datasets using the same fixed permutation for training and testing. As shown in Figure~\ref{fig:feasibility-b}, validation accuracy remains nearly identical across 10 random seeds. This demonstrates the \textit{semi}-permutation-invariance of neural networks: despite spatially distorted inputs, the model produces consistent computation results while benefiting from increased data privacy.


\vspace{1ex}

\noindent \textbf{Challenges and Core Idea.}
While training models on shuffled data can help preserve confidentiality, two challenges arise.  
First, models trained on shuffled data require shuffled inputs during both training and inference, adding computational overhead for resource‑constrained mobile and IoT devices.  
Second, widely used data augmentation techniques assume spatial coherence (e.g., random cropping, horizontal flipping)~\cite{perez2017effectiveness,shorten2019survey}, making them incompatible with shuffled data. Adapting augmentations to each permutation increases complexity and reduces generalizability.

To address these limitations, we leverage the interchangeability of shuffling \textit{data} and shuffling \textit{weights}, enabled by permutation invariance in neural networks. When trained on shuffled data, network weights implicitly incorporate the underlying permutation. By rearranging these weights, the model can behave as though it were trained on unshuffled data. Conversely, shuffling the weights of a model trained on unshuffled data enables equivalent behavior on shuffled inputs, thereby preserving confidentiality.

This weight‑shuffling strategy removes the need for repeated data shuffling during training and inference, reduces overhead, and restores compatibility with standard data augmentation. As a result, it simplifies system design while maintaining confidentiality and computation correctness.
\vspace{-2ex}

\section{Threat Model}
\label{sec:threat}

\noindent\textbf{Goals and Capability.} We address two key adversaries in federated learning: \textit{integrity attackers} (malicious clients) aiming to degrade global performance via data or model poisoning, and \textit{confidentiality attackers} (honest-but-curious server) attempting to infer private data from updates, potentially colluding with \textit{spy clients} to harvest auxiliary information.

\vspace{0.5ex}
\noindent\textbf{Assumptions.} We assume that attackers control local training but cannot interfere with server-side aggregation or benign clients, with malicious participants limited to $<50\%$~\cite{fang2020local}. 
Importantly, we assume that clients possess a general-purpose TEE, which is a practical reality on modern mobile computing platforms. TEEs (e.g., ARM TrustZone, Apple Secure Enclave) are ubiquitous hardware standards already powering essential daily services, from biometric authentication (e.g., FaceID) to mobile payments (e.g., Samsung Pay). Furthermore, recent advancements confirm that TEEs can efficiently handle computation-intensive AI workloads~\cite{mo2021ppfl,choi2022guardiann, mo2020darknetz}. Thus, leveraging them for lightweight shuffling in our proposed scheme ensures robust hardware-backed isolation against adversaries with negligible overhead.
\begin{figure*}[t!]
    \centering    
    \includegraphics[width=0.9\linewidth]{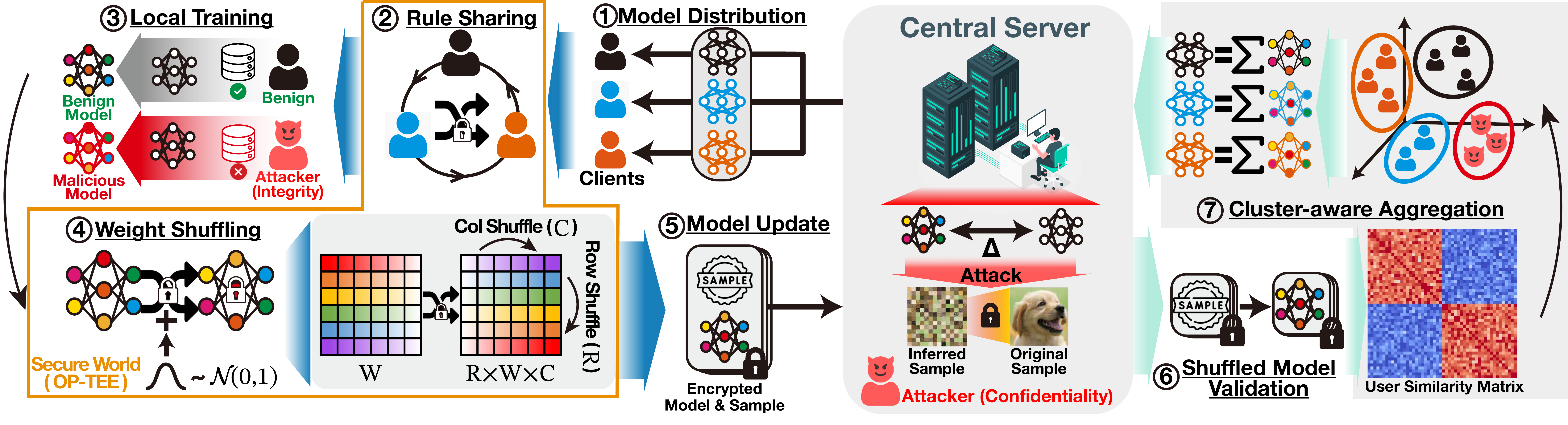}
    \vspace{-3ex}
    \caption{Overall workflow of \name{}.}
    \label{fig:overview}
    \vspace{-3ex}
\end{figure*}
\section{Framework Design}
\label{sec:design}

\subsection{Overview of \system}
\label{subsec:overview}

We now present \name{}, a secure and efficient federated learning framework that addresses both integrity and confidentiality attacks. \name{} leverages two core modules, \textit{weight shuffling} and \textit{shuffled model validation}, to exploit permutation equivariance and invariance properties of neural networks.

Figure~\ref{fig:overview} illustrates the overall workflow of \name{}:
\textbf{[Server-side]} \circled{1} The federated learning server distributes a global model to each client. 
\noindent \textbf{[Client-side]} \circled{2} Selected clients share a \textit{shuffling rule} (i.e., random seed for generating shuffling order) via secure p2p communication (Sec.~\ref{subsec:rs}). Note that this rule exchange is performed securely end-to-end between client TEE modules.
\circled{3} Participants of the federated learning round train the neural network with their locally collected data.
\circled{4} Upon completing local training, each client shuffles its model weights inside its TEE using the shared shuffling rule to ensure confidentiality (\textit{Weight Shuffling} - Sec.~\ref{subsec:weightshuffling}). Gaussian perturbation is then added to the shuffled weights before passing them to the clients' normal world for transmission.
\circled{5} Participating clients transmit shuffled parameters and a small number of input samples (as small as one) to the server. Note that input samples are also shuffled within the TEE to prevent confidentiality being compromised.

\noindent \textbf{[Server-side]} \circled{6} The server performs integrity check via \textit{shuffled model validation} by measuring the similarity of output vectors from the shuffled models and data (Sec.~\ref{subsec:intcheck}). The consistent shuffling of the model weights and data ensures valid outputs.
%
\circled{7} Finally, \name{} clusters the clients based on output similarity. Given that malicious and benign models exhibit different computation results, grouping the clients based on their similarity effectively assures that benign clients are not affected by malicious model updates.

To summarize, in Step \circled{4}, weight shuffling prevents the server (or spy client) from inferring client data by obscuring the model weights while preserving permutation invariance so that the vectors computed in Step \circled{6} remain valid. This allows the \textit{shuffled validation} process to detect poisoning attacks by clustering (and separating) malicious clients in Step \circled{7}. Moreover, since all clients share a common shuffling rule, model aggregation proceeds seamlessly.

\vspace{-2ex}
\subsection{Shuffling Rule Sharing}
\label{subsec:rs}

\name{} is designed under the condition that all clients share a common shuffling order rule. While we detail operations on how clients share this rule among themselves below, if the server somehow gains access to this rule, it could revert the shuffled models to their original state, posing a security risk.
To mitigate this, it is essential to establish an information asymmetry between the server and clients. \name{} leverages a secure communication channel between client Trusted Execution Environments (TEEs), ensuring the rule remains isolated within the secure world. Mechanisms such as secure multi-party computation~\cite{zhao2019secure}, zero-knowledge proofs~\cite{boo2021litezkp}, or peer-to-peer communication~\cite{castro1999practical} can ensure that neither the server nor a client's normal-world (potential spy client) gains access to the shuffling rule.

For shuffling rule exchange, \name{} employs a Practical Byzantine Fault Tolerant (PBFT) mechanism~\cite{castro1999practical}. 
Specifically, at the start of round $t$, the server designates a leader from the participant set $C_t$, prioritizing clients with stable connectivity (e.g., Wi-Fi) and power to ensure robustness. 
The leader then generates a random seed $\text{seed}_t$ within its TEE and initiates PBFT consensus among $C_t$. 
We assure that this process incurs negligible overhead, as it involves only active participants (typically 10-20 clients) exchanging a single 32-bit integer seed, avoiding global broadcasting or heavy payload transmissions.
Upon gathering $2f+1$ commit messages, a certificate attesting to the agreed seed is issued and forwarded to non-participating devices via authenticated channels. Should the leader fail, the standard PBFT view-change procedure elects a new leader. To address potential P2P failures (e.g., strict NAT), \name{} implements a \textit{Blind Relay} fallback, where the server forwards the TEE-encrypted rule without decryption capability.

Note that even with a malicious leader, PBFT ensures that consensus is achieved and the shuffling rule remains secure. Specifically, clients exchange the previous round's rule (for global model utilization) and the current round's new rule (for model updates). These rules are essentially lightweight seeds strictly isolated within TEE memory, preventing even OS-level adversaries from intercepting the shuffling logic.

\subsection{Weight Shuffling}
\label{subsec:weightshuffling}
Weight shuffling in \name{} prevents data leakage while preserving computational correctness by leveraging permutation equivariance and invariance. This module shuffles locally updated neural network weights before sharing them with the server. For clarity, we illustrate the process in the context of matrix multiplication, given its core role in neural network computations. Note that implementation details may vary depending on the baseline federated learning model architecture, as we discuss in Section~\ref{sec:implementation}.

In essence, weight shuffling in \system{} comprises two core operations: row shuffling and column shuffling. Consider matrix multiplication between an input matrix \(\mathbf{X} \in \mathbb{R}^{N \times d}\) and a weight matrix \(\mathbf{W} \in \mathbb{R}^{d \times h}\), where \(N\) is the number of vectors, \(d\) the input dimension, and \(h\) the embedding dimension. To shuffle the weight rows, a row-permutation matrix \(\mathbf{R}\) is applied to \(\mathbf{W}\) (i.e., \(\mathbf{R} \times \mathbf{W}\)). This operation is equivalent to shuffling the input, thereby ensuring data confidentiality. Note that a permutation matrix \(\mathbf{R}\) satisfies \(\mathbf{R}^{\top}\mathbf{R}=\mathbf{I}\), where \(\mathbf{I}\) is the identity matrix. Thus, applying the same permutation to \(\mathbf{X}\) via \(\mathbf{R}^{\top}\) (i.e., \(\mathbf{X}\times \mathbf{R}^{\top}\)) keeps the computation unchanged, since $(\mathbf{X} \mathbf{R}^{\top})(\mathbf{R}\mathbf{W})=\mathbf{XIW}=\mathbf{XW}$ (c.f., Figure~\ref{fig:perm_vis} (b)).

Weights can be further shuffled along the columns by multiplying the weight matrix \(\mathbf{W}\) with a column-permutation matrix \(\mathbf{C}\). Since column shuffling only alters the output order and not the computation itself (c.f., Figure~\ref{fig:perm_vis} (c)), no additional permutation is needed as input. However, for the bias term \(\mathbf{b} \in \mathbb{R}^{h}\) in a linear layer (i.e., \(\mathbf{X} \times \mathbf{W} + \mathbf{b}\)), the same column permutation must be applied to \(\mathbf{b}\) to maintain alignment with shuffled weights. Moreover, since the output of one layer becomes the input of the next, the column shuffling order of the current layer can be directly inherited as the row shuffling order for the weights of the subsequent layer.

This shuffling operation for matrix multiplication (i.e., linear layers) leverages permutation invariance and equivariance, the foundations of most neural network computations. However, uniformly applying this to all weights may not work as intended because model behaviors and designs vary. We elaborate on adapting the shuffling operation for various advanced neural network architectures in Section~\ref{sec:implementation}.


In operation, weight shuffling occurs strictly within the clients' TEE. Once local training completes, model weights are transferred to the secure world for shuffling, ensuring that the shuffling rule remains completely hidden from the normal world, where spy clients or compromised operating systems might attempt leakage. However, considering the strict memory constraints of commodity TEEs (e.g., limited secure RAM~\cite{choi2022guardiann}), loading the entire model as a whole is often infeasible. To address this, \name{} employs a \textit{layer-wise pipelining} mechanism. Instead of a bulk transfer, weights are loaded into the secure world in sequential chunks (i.e., per layer). Within the TEE, each chunk is shuffled and immediately offloaded to the normal world, ensuring that the secure memory footprint remains minimal and constant regardless of the total model size. For additional protection, \name{} adds small Gaussian perturbations to the shuffled weights, preventing attackers from reconstructing the shuffling rule by comparing pre- and post-shuffled weights. Furthermore, recognizing that random number generation can be computationally expensive within TEEs, \name{} pre-generates these noise vectors prior to the shuffling phase. This combination of pipelined execution and pre-computation ensures that \name{} maintains robust security with high latency efficiency.

\subsection{Integrity Check}
\label{subsec:intcheck}

Integrity check in \name{} is designed to identify and isolate malicious models from benign ones through a two-step process: \textit{shuffled model validation} and \textit{cluster-aware aggregation}. 

\subsubsection{Shuffled Model Validation}
\label{subsubsec:sv}




In the Shuffled Model Validation module, the server exploits a small number of client-provided shuffled samples as inputs to the uploaded models to validate the integrity of the gathered models. Despite the sample obfuscation, the corresponding weight shuffling preserves computational validity; thus, the server should be able to trust the outputs if the same shuffle rules are applied. If an attacker uploads models with mismatched shuffling orders for data and weights, the resulting outputs will be inconsistent, distinguishing these models from legitimate ones and forcing attackers to comply with the correct rules.


Note that malicious models produce outputs that deviate from those of benign models, revealing distinctive patterns. \name{} identifies such deviations by computing pairwise cosine similarity among the output vectors of all models~\cite{park2023attfl}. Malicious models typically exhibit lower cosine similarity with benign ones, providing a basis for their classification and isolation. Clients are then clustered based on pairwise cosine similarities, with the similarity matrix (\(S\)) transformed into a distance matrix (\(1-S\)) to interpret similarity as a distance metric, which is used as input to the DBSCAN algorithm~\cite{ester1996density}. \name{} leverages DBSCAN given its ability to identify outliers and its independence from predefined cluster numbers, making it well-suited for federated learning scenarios where the presence of attackers is unknown.

\subsubsection{Cluster-aware Aggregation}
\label{subsubsec:caa}

Finally, we implement cluster-aware aggregation to separate the malicious models. Algorithm~\ref{alg:cluster_aware_aggregation} shows the pseudo-code for this process. In each federated learning round, the cluster labels computed by DBSCAN serve as \textit{pseudo-labels} that capture per-round client similarities; with independent labels across rounds (e.g., labels from round \(t\) do not carry over to round \(t+1\)). To ensure consistency, \name{} assigns global cluster labels by leveraging clustering histories for effective cross-round clustering. Clients are categorized into two groups: (1) new participants without a global cluster, and (2) returning clients with an existing global cluster (see lines 3 in Algorithm~\ref{alg:cluster_aware_aggregation}).


\begin{algorithm}[t]
\caption{Cluster-Aware Aggregation in \name{}}
\label{alg:cluster_aware_aggregation}
\KwIn{Client models, pseudo-cluster labels for round \(t\), global clusters \(G\)}
\KwOut{Updated global clusters and models}

Initialize $G_{\text{new}} \gets G$\;

\ForEach{pseudo-cluster $C_p$ in round $t$}{
    Partition clients: $C_{\text{new}}$ (no global label), $C_{\text{existing}}$ (with global label)\;
    
    \uIf{$C_{\text{existing}} = \emptyset$}{
        Assign global cluster $g_{\text{new}}$ to $C_p$, update $G_{\text{new}}$\;
    }
    \Else{
        Assign dominant global label \(g_{\text{dom}}\) to $C_p$\;
        \If{multiple global labels exist in $C_p$}{
            Merge and update $G_{\text{new}}$\;
        }
    }
}
\ForEach{$g \in G_{\text{new}}$}{Aggregate models for clients in $g$\; }
\Return $G_{\text{new}}$, aggregated models\;
\end{algorithm}

Upon performing shuffled model validation, if all clients in a cluster are first-time participants without global cluster assignment, \name{} creates and assigns new global clusters (lines 4-5). Conversely, if a pseudo-cluster contains clients with pre-existing global clusters, all clients in the cluster are assigned the global label of those clients (lines 6-7).


Furthermore, if clients previously belonging to different global clusters are grouped in the same pseudo-cluster during the current round, \name{} merges their global clusters. Rather than merging all clients indiscriminately, \name{} adopts a conservative approach by reassigning only the overlapping clients to the new cluster (lines 8-9). This strategy ensures that a single attacker infiltrating a benign cluster cannot force all attackers into the same benign group.

Once global clusters are assigned, \name{} aggregates models within each cluster, producing an updated model for each cluster (lines 10-11). Although we currently use averaging, alternative aggregation schemes can be used, offering design flexibility. Aggregating models solely within the same cluster prevents compromised models from contaminating the global model shared across clusters. Note that, while the server is unaware of the shuffling rule, it can seamlessly aggregate parameters since they exploit consistent rules (c.f., Sec.~\ref{subsec:weightshuffling}). Furthermore, we note once more that the impact of the added Gaussian perturbation is minimal~\cite{wu2022noisytune}.

This approach maintains the integrity of federated learning by ensuring that malicious models cannot propagate their impact to benign clusters~\cite{liu2021distfl}. Moreover, by leveraging shuffled model validation, \name{} achieves robust defense without exposing or interpreting clients' data, thereby preserving both system security and data confidentiality. Overall, this dual protection reinforces \name{}'s effectiveness in safeguarding federated learning against security threats.

\section{Implementation}
\label{sec:implementation}

As previously noted, the implementations of the weight shuffling operation can vary for different model architectures. In this section, we provide a detailed explanation of the weight shuffling process for representative architectures.


\vspace{1ex}
\noindent\textbf{Multi-layer Perceptron (MLP).} As a simple architecture, we demonstrate the weight shuffling operation in an MLP. In an MLP with \( L \) layers, the output \( y \) is computed as follows:
\begin{equation}
\begin{split}   
h^{(l)} &= \sigma(h^{(l-1)}W^{(l)\top} + b^{(l)}), \quad l = 1, \dots, L-1, \\
h^{(0)} &= x,\;\;\; y = h^{(L-1)}W^{(L)\top} + b^{(L)},
\end{split}
\end{equation}
Here, \( h^{(l)} \) denotes the activations of the \( l \)-th layer, \( W^{(l)} \) the weight matrix associated with the \( l \)-th layer, \( b^{(l)} \) is the bias vector for the \( l \)-th layer, and \( \sigma(\cdot) \) is a non-linear activation function such as ReLU, sigmoid, or tanh.

A critical observation is that commonly used activation functions are parameter-free and permutation-equivariant, meaning their behavior remains unaffected by input shuffling. Therefore, activation functions can be excluded from weight shuffling considerations. This leaves the linear transformations at each layer, represented as matrix multiplications (i.e., \(h^{(l-1)}W^{(l)\top}\)), as the primary focus. The weight matrices and bias vectors in these transformations can be shuffled following the strategies outlined earlier, ensuring that computational outcomes remain consistent. By iteratively applying this mechanism across layers, the entire network retains compatibility with weight shuffling, preserving both computational integrity and model confidentiality.


\begin{figure}[t!]
    \centering
    \includegraphics[width=0.9\linewidth]{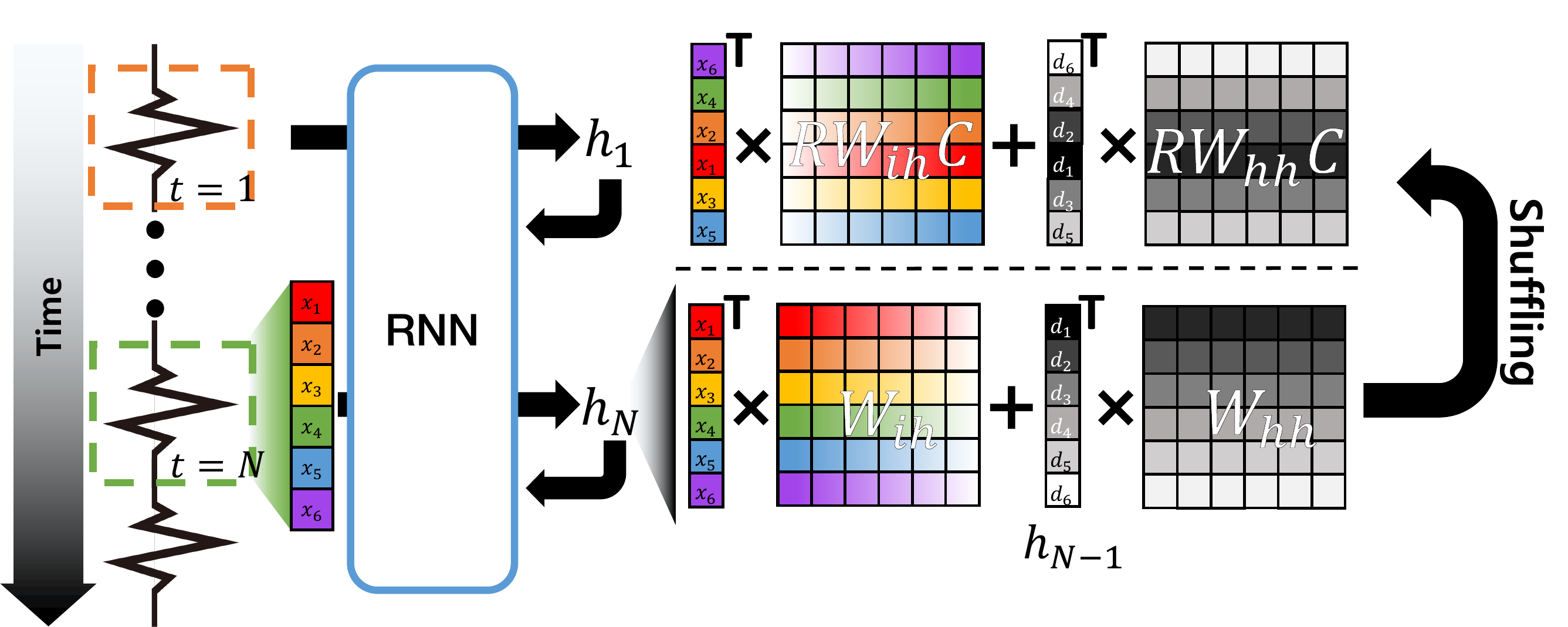}
    \vspace{-3ex}
    \caption{Weight shuffling on a sequential model.}
    \vspace{-2ex}
    \label{fig:rnn}
\end{figure}

\vspace{1ex}
\noindent \textbf{Sequential Models (RNN and LSTM).} For sequential models such as RNNs and LSTMs, the weight shuffling process is similar to that of MLPs, with the key addition of handling hidden state operations that capture temporal dependencies. Figure~\ref{fig:rnn} shows an RNN where the next output is computed by combining the current input (\(x_t\), \(W_{ih}\)) with the previous hidden state (\(h_{t-1}\), \(W_{hh}\)). Despite this complexity, the core computations are still based on matrix multiplications; thus, exhibiting permutation equivariance and invariance, allowing weight shuffling without compromising correctness.

Note that this property extends to other sequential architectures such as LSTMs and GRUs, as they similarly rely on matrix operations for processing inputs and hidden states, making them compatible with weight shuffling.

\begin{figure}[t!]
    \centering
    \includegraphics[width=0.96\linewidth]{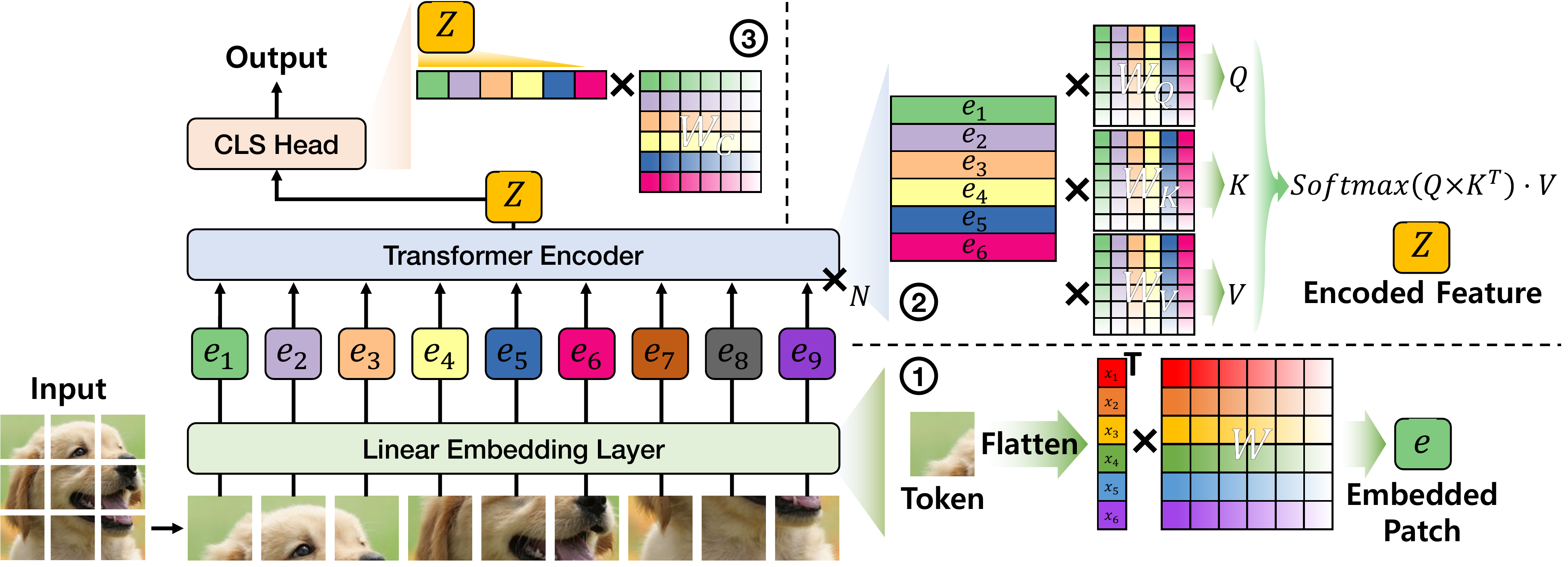}
    \vspace{-3ex}
    \caption{Workflow with a transformer baseline.}
    \label{fig:vit}
\end{figure}

\vspace{0.05in}

\noindent \textbf{Transformers.} As a widely used modern neural network architecture, transformers operates in three main steps~\cite{vaswani2017attention}, as depicted in Figure~\ref{fig:vit}. First, the input is segmented into a sequence of tokens, which are passed through a linear embedding layer to generate token embeddings. These embeddings are then processed by multiple encoder layers that analyze the input. Each encoder employs a multi-head attention mechanism comprising three learnable matrices: \( W_{Q} \) (query), \( W_{K} \) (key), and \( W_{V} \) (value). Finally, the encoded feature representation, \( Z \), is fed into the classification head (i.e., \texttt{CLS Head}) to produce the final output. Importantly, the encoded feature \( Z \) preserves the input tokens order, ensuring the model output's alignment with the input sequence.


From a data perspective, shuffling can be applied at two levels: \textit{intra-token} and \textit{inter-token}. At the intra-token level, individual elements within a token are permuted. At the inter-token level, the sequence of tokens is rearranged. Since all tokens pass through a linear embedding layer—essentially a matrix multiplication—the permutation invariance property ensures that identical computations are achieved by shuffling the embedding weights accordingly.

For inter-token shuffling, the sequential order of tokens is preserved for the attention mechanism (e.g., row order \(e_{1}, e_{2}, \dots, e_{n}\)), where positional features are learned. In the final \texttt{CLS Head}, these features determine the predictions. Thus, by appropriately shuffling classification head weights, consistent outputs are kept even with altered token orders.



We note that operations beyond matrix multiplications also exhibit either permutation equivariance (e.g., softmax) or permutation invariance (e.g., mean and variance computations in layer norm.). Consequently, these operations are unaffected by weight shuffling, ensuring that results remain consistent. This consistency allows \name{} to uphold both the confidentiality of client data and the integrity of the system.
\section{Evaluation}
\label{sec:eval}
Now we present evaluation results of \name{} from extensive experiments using four datasets and various comparison baselines. We exploit \system{} implementations for various baseline models suitable for each dataset in our evaluations. 
\subsection{Experiment Setup}
\label{subsec:setup}

\vspace{1ex}

\noindent\textbf{Dataset and Model.}
We use four datasets suitable for mobile federated learning scenarios as detailed below.

\noindent$\bullet$ \textbf{MNIST}~\cite{deng2012mnist}: 60K train / 10K test grayscale digits (28$\times$28). Model used: 3-layer MLP~\cite{rosenblatt1958perceptron}.

\noindent$\bullet$ \textbf{CIFAR10}~\cite{krizhevsky2009learning}: 60K color images (32$\times$32, 10 classes). Model used: MLP-Mixer~\cite{tolstikhin2021mlp}.

\noindent$\bullet$ \textbf{STL10}~\cite{coates2011analysis}: 13K natural images (96$\times$96). Model used: Vision Transformer~\cite{dosovitskiy2020image}.

\noindent$\bullet$ \textbf{MIT-BIH}~\cite{moody1992bih}: 70K 5sec ECG segments. Model used: 3-class RNN~\cite{rumelhart1986learning}. Show time-series data applicability~\cite{park2023self, lee2024gmt}.





\vspace{1ex}
\noindent\textbf{Baselines.} For confidentiality attacks, we evaluate three baseline defense mechanisms: no defense, gradient pruning~\cite{zhu2019deep}, and differential privacy~\cite{abadi2016deep, dwork2014algorithmic}. The gradient pruning approach zeros out small-amplitude gradients to reduce data leakage but requires a carefully balanced pruning ratio; higher ratios mitigate leakage but risk discarding useful knowledge. Differential privacy (DPSGD~\cite{abadi2016deep}), adds gradient perturbation to obscure sensitive information, with perturbation levels as small, medium, and large~\cite{abadi2016deep}. Similar to gradient pruning, differential privacy requires careful perturbation level tuning to balance between accuracy and privacy. We exclude comparisons with heavy cryptographic protocols such as Homomorphic Encryption (HE), as their prohibitive computational overhead induced on mobile devices renders them impractical for real-time use~\cite{jiang2025towards} compared to the lightweight design of \name{}.

For integrity attacks, we use FedAvg~\cite{mcmahan2017communication} as a baseline, being a security-less naive aggregation scenario. We also evaluate robust methods leveraging statistical priors, including Median and Trimmed Mean~\cite{yin2018byzantine}. The Median approach replaces averaging with the median to mitigate the influence of extreme outliers, while the Trimmed Mean approach excludes outliers before averaging. Additionally, we compare \name{} with MultiKrum~\cite{blanchard2017machine} and FLTrust~\cite{cao2020fltrust}, which represent similarity-based approaches. MultiKrum selects updates with the smallest pairwise Euclidean distances after discarding the most distant ones, and FLTrust assigns trust scores based on cosine similarity with a server-trained reference model, adjusting aggregation weights to suppress malicious updates.

\vspace{1ex}
\noindent\textbf{Attack Methods.} For confidentiality attacks, we use the inverting gradient attack~\cite{geiping2020inverting} for 100 test samples as baseline. This approach reconstructs local data by optimizing inputs to produce gradients similar to those from the client, achieving high performance with minimal additional assumptions~\cite{huang2021evaluating}.

For integrity attacks, we evaluate three poisoning methods: label flipping, noise injection, and backdoor attacks~\cite{fang2020local}. In label flipping, a subset of training data labels are altered to mislead the local model. Noise injection introduces random noise into model parameters prior to transmission, aiming to degrade the global model performance. Backdoor attacks embed specific triggers in the training data, causing the model to behave maliciously when exposed to those triggers.

\vspace{1ex}
\noindent\textbf{Misc. Configurations.} Unless specified, datasets are distributed across 100 clients (25 attackers), with data following Dirichlet distribution with $\alpha$=0.5~\cite{hsu2019measuring}. In each federated learning round, 10 participants are selected, and models are locally trained for 5 epochs using Adam optimizer~\cite{kingma2014adam} with a learning rate 1e-3 and batch size 64. For server-side validation, we use one random shuffled sample from each client. Besides evaluations on real embedded platforms in Section~\ref{sec:real}, all experiments ran on a server with an NVIDIA RTX 2080 GPU, an Intel Xeon Silver 4210 CPU 2.20GHz, and 64GB RAM.

\subsection{Overall Performance}
\begin{figure}[t!]
    \centering
    \includegraphics[width=0.9\linewidth]{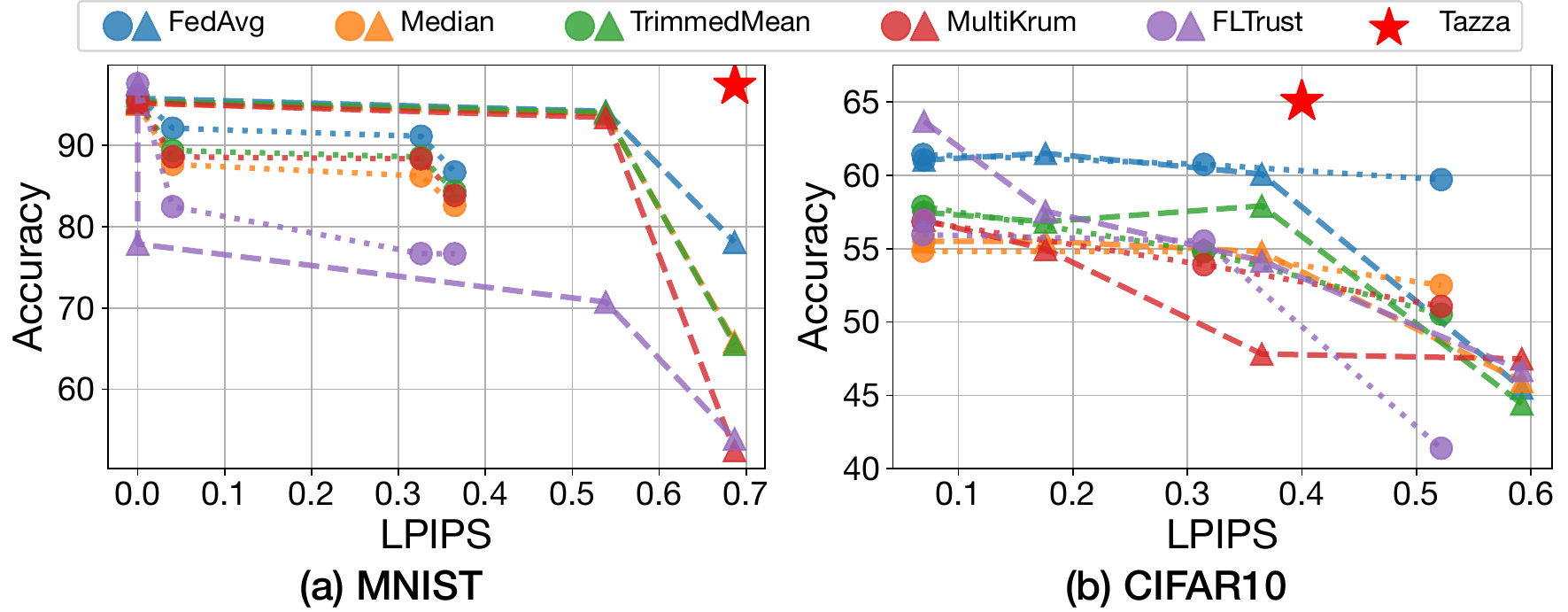}
    \vspace{-2ex}
    \caption{LPIPS and accuracy of varying federated learning configurations. $\blacktriangle$ and $\bullet$ denote applying gradient pruning and DPSGD, respectively.}
    \vspace{-3ex}
    \label{fig:overall}
\end{figure}

To evaluate the overall performance of \name{}, we conducted experiments using the MNIST-MLP and CIFAR10-Mixer configurations with 75 benign clients and 25 label-flipping (integrity) attackers. To understand \name{}'s confidentiality-preserving capabilities, the server executed an inverting gradient attack. The attack effectiveness was measured using Learned Perceptual Image Patch Similarity (LPIPS)~\cite{zhang2018unreasonable}, which quantifies semantic distance in image data. Lower LPIPS indicate greater similarity between the original and reconstructed samples, signifying higher attack success.

Figure~\ref{fig:overall} presents the global model accuracy under label-flipping attacks and LPIPS across integrity and confidentiality defense schemes. Since each cluster in \name{} has distinct model parameters, we also report the average accuracy of all benign clients' models on the test dataset. The legend indicates different integrity defense schemes, with triangle (\(\blacktriangle\)) and circle (\(\bullet\)) markers representing confidentiality defense schemes using gradient pruning and DPSGD, respectively. Gradient pruning is applied with pruning ratios of 70\%, 90\%, and 99\%, while DPSGD set small, medium, and large perturbation levels, as in prior work~\cite{abadi2016deep}. Each scheme includes three plots: the leftmost plot corresponds to the weakest confidentiality defense configuration (70\% pruning for gradient pruning and small perturbation for DPSGD), while the rightmost plot shows the strongest configuration (99\% pruning for gradient pruning and large perturbation for DPSGD).

Existing confidentiality defense mechanisms improve privacy with higher LPIPS, but often with accuracy drops. In contrast, \name{} achieves both high LPIPS and accuracy, being robust against integrity and confidentiality attacks.

\subsection{Confidentiality Attack Mitigation}
\begin{figure}[t!]
    \centering
    \includegraphics[width=0.9\linewidth]{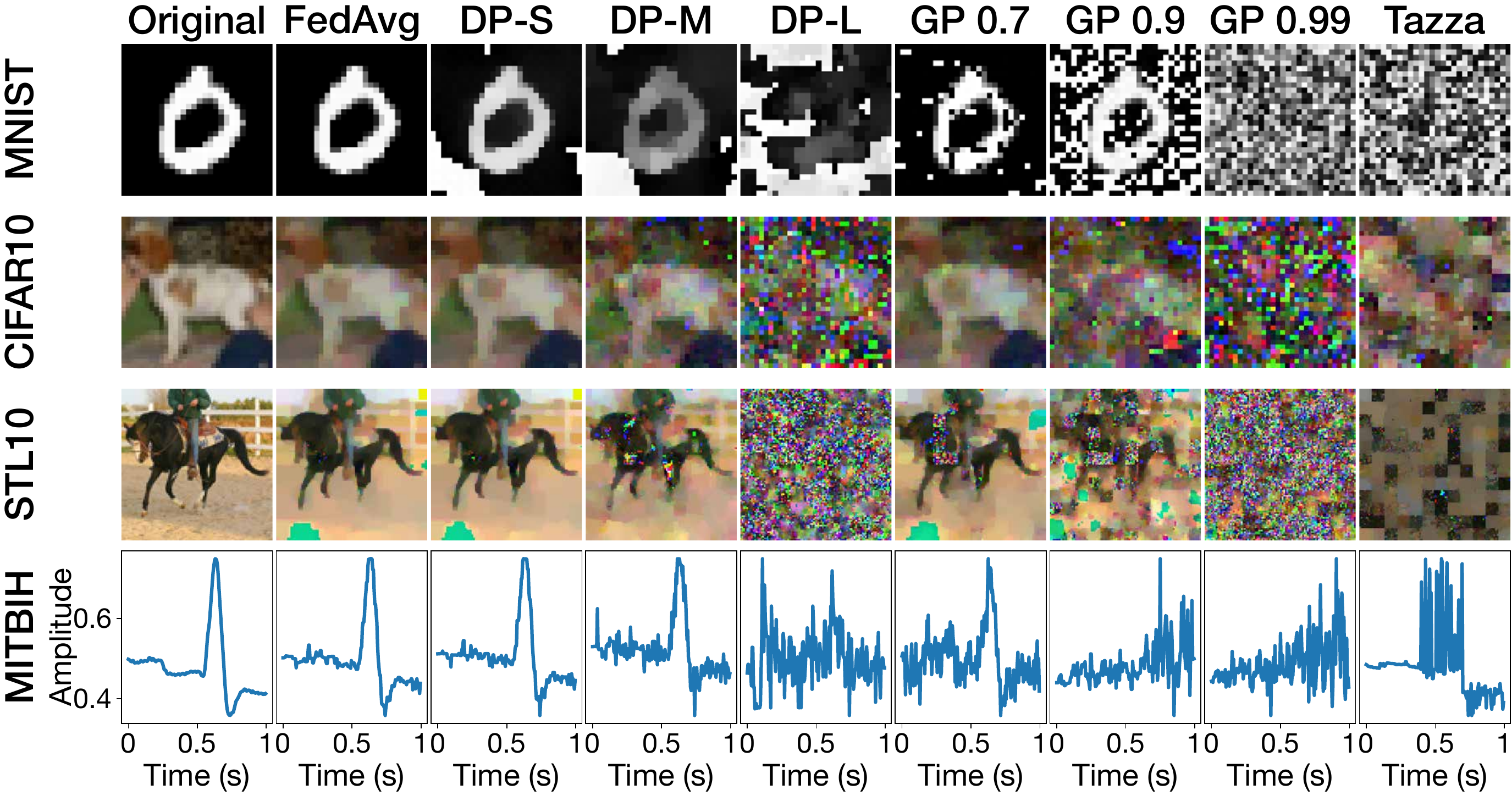}
    \vspace{-3ex}
    \caption{Sample confidentiality attack results for different dataset/defense configurations.}
    \vspace{-3ex}
    \label{fig:ca-samples}
\end{figure}
\begin{table}[t]
\begin{adjustbox}{width=0.9\linewidth, center}
\begin{tabular}{|c|l|c|c|c|c|}
\hline
\multicolumn{1}{|l|}{}                    &         & FedAvg      & GP 0.9              & DP-M        & \textbf{Tazza}               \\ \hline
\multirow{4}{*}{Avg PSNR ($\downarrow$)}  & MNIST   & 47.08±12.33 & \textbf{4.92±0.53}  & 6.88±2.17   & 4.95±0.21           \\ \cline{2-6} 
                                          & CIFAR10 & 18.11±6.77  & 13.17±3.73          & 15.60±4.01  & \textbf{11.74±2.92} \\ \cline{2-6} 
                                          & STL10   & 12.21±4.88  & 10.23±2.33          & 12.14±4.67  & \textbf{9.91±2.38}  \\ \cline{2-6} 
                                          & MITBIH  & 23.56±11.09 & \textbf{-7.37±3.78} & 19.33±13.39 & 17.01±8.74          \\ \hline
\multirow{4}{*}{Best PSNR ($\downarrow$)} & MNIST   & 61.28       & 6.34                & 14.77       & \textbf{5.31}       \\ \cline{2-6} 
                                          & CIFAR10 & 30.46       & \textbf{20.47}      & 24.33       & 21.12               \\ \cline{2-6} 
                                          & STL10   & 21.32       & \textbf{15.15}      & 20.08       & 16.18               \\ \cline{2-6} 
                                          & MITBIH  & 35.35       & \textbf{0.08}       & 32.34       & 22.69               \\ \hline
\multirow{3}{*}{Avg LPIPS ($\uparrow$)}   & MNIST   & 0.04±0.16   & 0.69±0.04           & 0.52±0.09   & \textbf{0.75±0.02}  \\ \cline{2-6} 
                                          & CIFAR10 & 0.36±0.24   & 0.61±0.10           & 0.51±0.12   & \textbf{0.62±0.07}  \\ \cline{2-6} 
                                          & STL10   & 0.63±0.17   & 0.73±0.07           & 0.68±0.12   & \textbf{0.75±0.05}  \\ \hline
\multirow{3}{*}{Best LPIPS ($\uparrow$)}  & MNIST   & 0.00        & 0.54                & 0.33        & \textbf{0.69}       \\ \cline{2-6} 
                                          & CIFAR10 & 0.07        & 0.37                & 0.31        & \textbf{0.40}       \\ \cline{2-6} 
                                          & STL10   & 0.34        & \textbf{0.62}       & 0.50        & 0.58                \\ \hline
\end{tabular}
\end{adjustbox}
\caption{Average and best PSNR/LPIPS between original and inferred samples from confidentiality attack.}
\vspace{-3ex}
\label{tab:ca-result}
\end{table}

We evaluate the efficacy of \name{} in mitigating confidentiality attacks using the PSNR and LPIPS between the original training samples and those inferred from the attack. PSNR assesses structural similarity, while LPIPS evaluates semantic distance for image data. Note that, as LPIPS is specific to images, only PSNR is reported for the MIT-BIH dataset. Table~\ref{tab:ca-result} presents the average and best PSNR and LPIPS for various dataset and defense scheme configurations. The best values represent the most similar samples among the 100 used in the experiments. Higher PSNR and lower LPIPS indicate greater similarity between inferred and original samples.

As Table~\ref{tab:ca-result} shows, FedAvg exhibits high PSNR and low LPIPS, indicating substantial local data leakage. In contrast, defenses such as gradient pruning with 90\% pruning (GP 0.9), DPSGD with medium perturbation (DP-M), and \name{} demonstrate competitive reductions in PSNR with LPIPS enhancement, effectively mitigating confidentiality attacks. While gradient pruning and differential privacy face challenges in balancing accuracy with confidentiality, \name{} successfully defends attacks without accuracy loss.

Figure~\ref{fig:ca-samples} presents the server-inferred training samples generated by different gradient inversion attacks. In the absence of a defense mechanism (FedAvg), local training data are completely exposed; significantly compromising the confidentiality of federated learning systems. Methods such as DPSGD with large perturbation (DP-L), pruning 99$\%$ of the local gradient effectively mitigate the gradient inversion attack, preserving the privacy of training data while sacrificing model training performance as noted earlier. On the other hand, when employing relatively weak defense mechanisms such as gradient pruning with pruning ratios 70$\%$ and 90$\%$ (GP 0.7, 0.9) or DPSGD with smaller perturbations (DP-S, M), the inferred samples reveal interpretable semantic information, though with reduced clarity, suggesting the need for stronger privacy-preserving techniques. As \name{}'s samples in Figure~\ref{fig:ca-samples} show, our proposed approach effectively conceals semantic information for both STL10-ViT and MIT-BIH-RNN configurations, suggesting its potential usage in high-dimension data and time-series data-based applications.


\subsection{Integrity Attack Mitigation}
\begin{figure}[t!]
    \centering
    \includegraphics[width=0.87\linewidth]{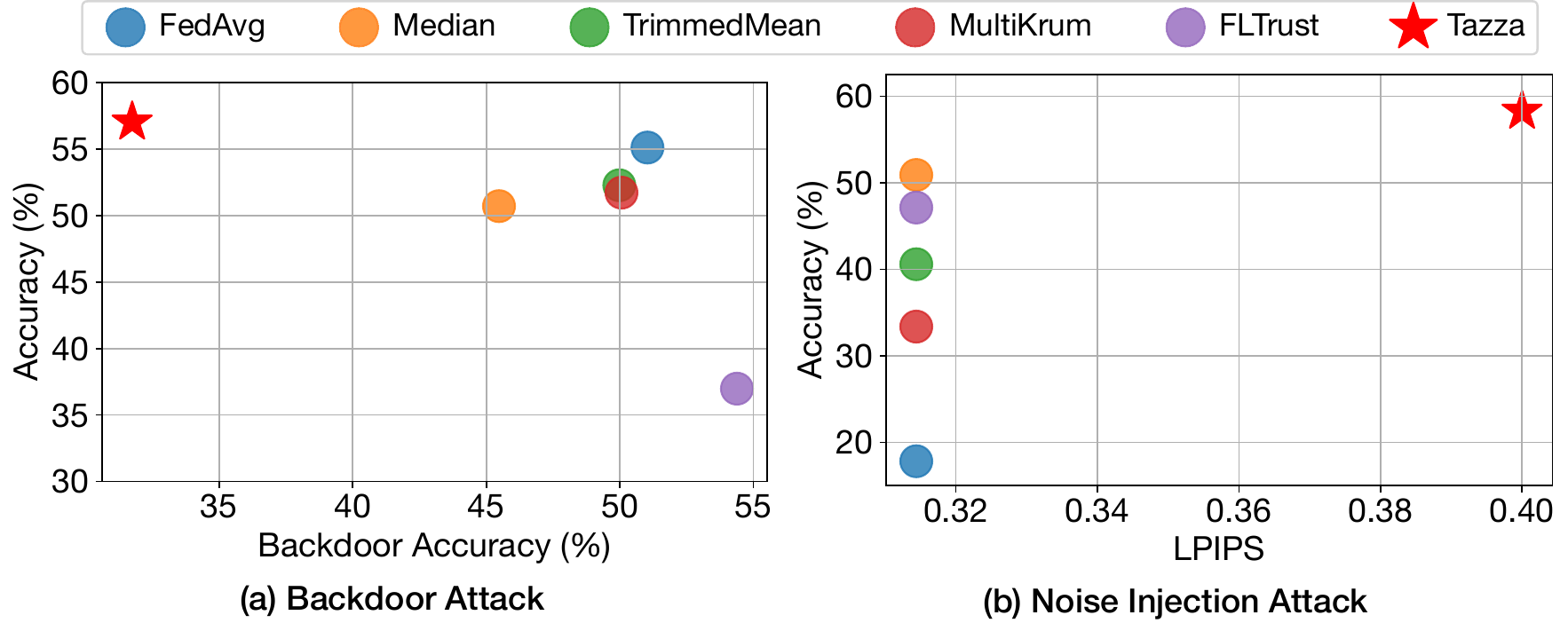}
    \caption{(a) Backdoor accuracy and global model accuracy for different defense schemes against backdoor attack. (b) LPIPS and accuracy of different defense mechanisms under noise injection attack.}
    \vspace{-1ex}
    \label{fig:ba}
\end{figure}

Next, we evaluate \name{}'s robustness against various integrity attacks, specifically focusing on backdoor and noise injection attacks. All baselines use the CIFAR10-Mixer configuration with medium-level DPSGD for confidentiality defense. For the backdoor attack, attackers assign specific labels to samples by injecting a 4$\times$4 red square patch into the upper-left corner of 50\% of their training data as the backdoor trigger~\cite{lee2024detrigger}. For the noise injection attack, we inject Gaussian noise ($\sim\mathcal{N}(0,1)$) with a scaling factor of 0.25 to model weights before uploading to the server.

Figure~\ref{fig:ba} (a) presents the global model accuracy (higher y-axis indicates better performance on normal samples) and backdoor accuracy (lower x-axis values indicate better attack mitigation for triggered samples). As shown, baseline methods struggle to counter backdoor attacks and often sacrifice global accuracy. On the other hand, \name{} achieves a 19.39\% reduction in backdoor accuracy compared to FedAvg while maintaining a global model accuracy of 57.03\%. Figure~\ref{fig:ba} (b) plots the average LPIPS and model accuracy under a noise injection attack. Similar to the backdoor attack case, \name{} shows high accuracy while mitigating the impact of the confidentiality attack (i.e., high LPIPS).

\begin{figure}[t!]
    \centering
    \includegraphics[width=0.9\linewidth]{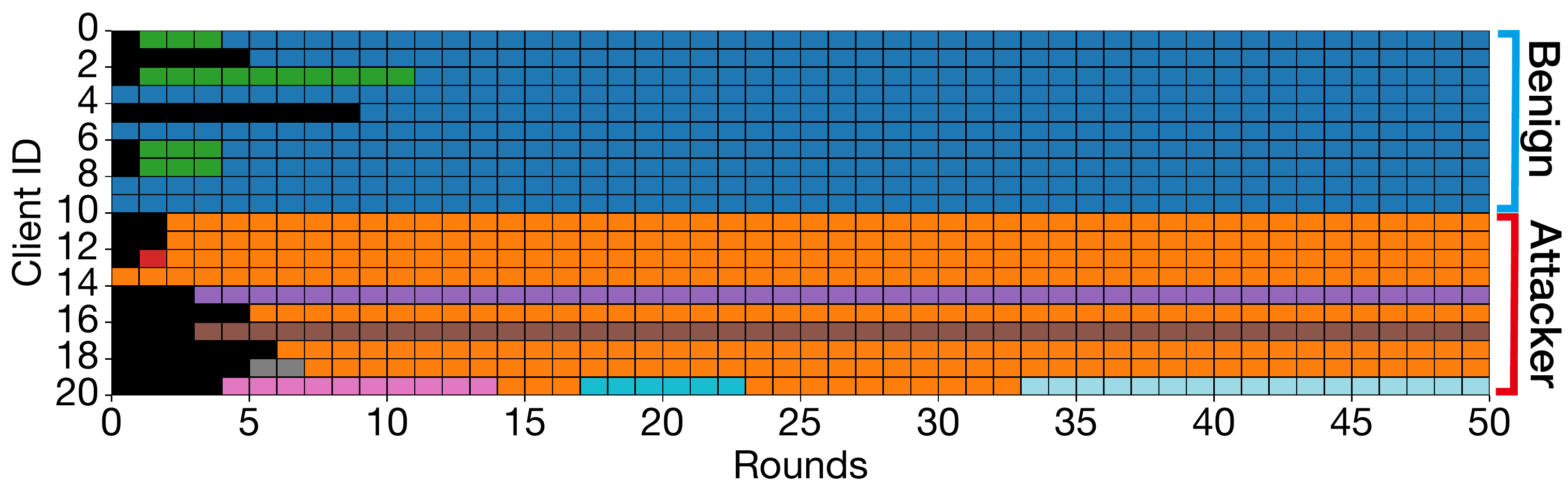}
    \vspace{-3ex}
    \caption{Visualization of client clustering. Different colors denote different clusters. Best viewed in color.}
    \vspace{-3ex}
    \label{fig:clustering}
\end{figure}

To validate \name{}'s capability to distinguish malicious behavior from natural data heterogeneity (i.e., non-i.i.d), we visualize cluster allocations for 10 benign and 10 backdoor clients in Figure~\ref{fig:clustering}. As shown, \name{} consistently isolates attackers (Clients 11-20) from benign ones (Clients 1-10). This empirical evidence confirms that the structural deviations induced by attacks are distinct from legitimate non-i.i.d variations, ensuring that \name{} precisely targets threats without misclassifying benign outliers.

\subsection{Real Embedded Platform Evaluation}
\label{sec:real}
\begin{figure}[t!]
    \centering
    \includegraphics[width=0.87\linewidth]{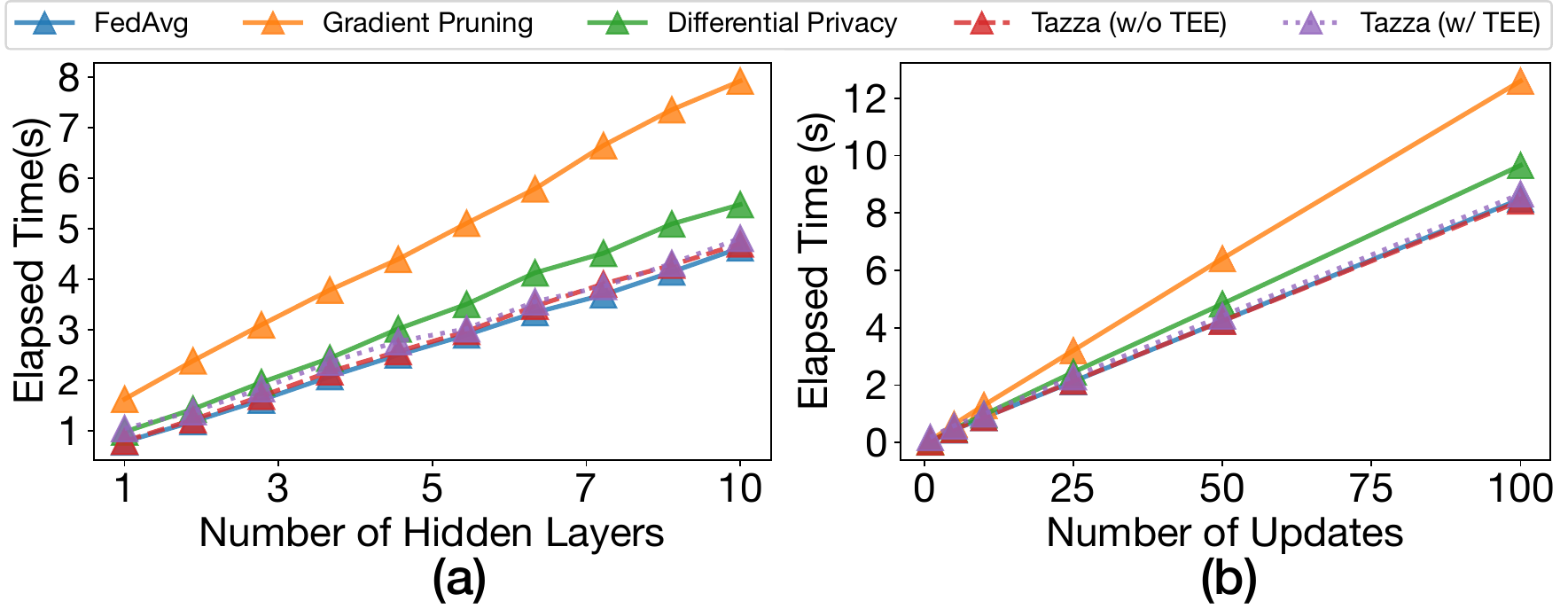}
    \vspace{-3ex}
    \caption{Latency of different confidentiality attack mitigation schemes with varying (a) hidden layer counts and (b) the number of local training updates.}
    \vspace{-1ex}
    \label{fig:latency}
\end{figure}


We now validate the performance of \name{} using four different embedded platforms: Raspberry Pi 3B~\cite{rb3b2023}, Nvidia Jetson Nano~\cite{nvidiaJetsonNano}, Nvidia Jetson TX2~\cite{nvidiaJetsonTx2}, and Nvidia Xavier NX~\cite{nvidiaxavier}. These experiments focus on the computational latency of executing \name{} compared to widely used attack mitigation schemes. Due to the instability of official TEE support, experiments on the Raspberry Pi, Nvidia Nano, and TX2 were conducted entirely in the normal world. For Xavier NX, we include experiments using OP-TEE~\cite{mo2020darknetz} implementations alongside the non-TEE configuration.

Figure~\ref{fig:latency} plots the computational latency of various defense mechanisms across different network depths (Figure~\ref{fig:latency}~(a)) and numbers of mini-batch updates during local training (Figure~\ref{fig:latency}~(b)) using the MNIST-MLP configuration on the Xavier NX platform (with and without TEE).
In Figure~\ref{fig:latency}~(a), the latency increases with neural network depth, exhibiting a near-linear trend across all methods. \name{} (with and without TEE) demonstrates a less-steep increase compared to alternative schemes such as gradient pruning and differential privacy. Figure~\ref{fig:latency}~(b) shows a similar trend when increasing the number of mini-batch updates. 
Note that using a TEE for weight shuffling and noise addition protects the system from spy clients that could potentially leak the shuffling rule to the server. This last-mile protection introduces a small amount of latency due to the (relatively) slower nature of secure-world computations~\cite{choi2022guardiann}. Nevertheless, given its light-weight nature, \name{} shows efficient latency performance even when exploiting TEE operations for weight shuffling.


\begin{figure}[t!]
    \centering
    \includegraphics[width=0.9\linewidth]{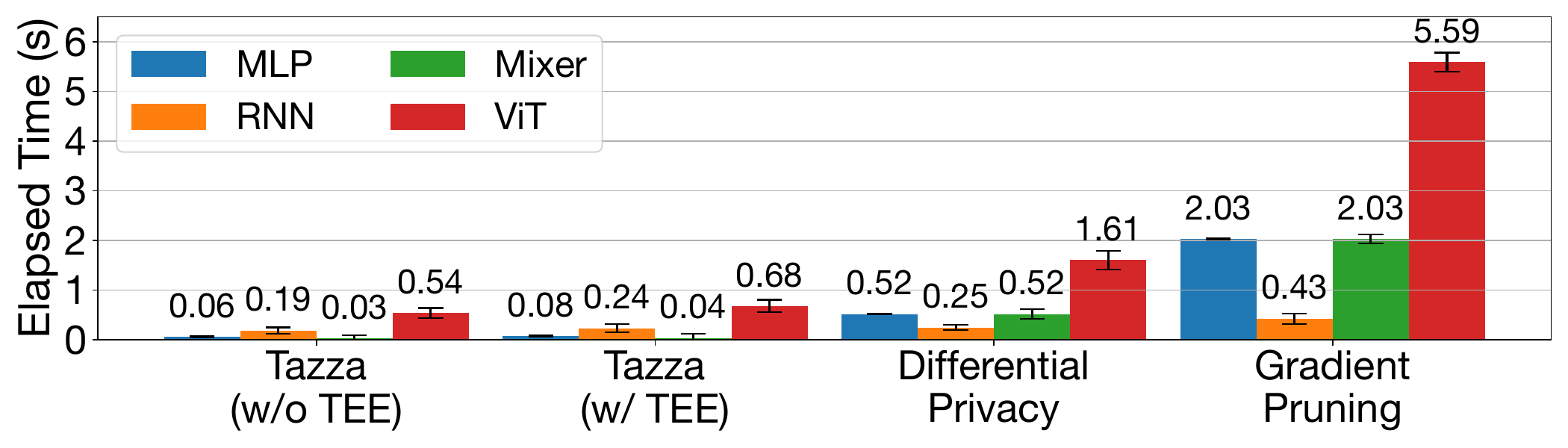}
    \vspace{-3ex}
    \caption{Latency of different confidentiality attack mitigation schemes with varying model architectures.}
    \vspace{-3ex}
    \label{fig:latency-model}
\end{figure}

Figure~\ref{fig:latency-model} plots the latency of core defense operations for each defense, excluding the model training latency and isolating only the overhead introduced by the confidentiality attack mitigation operations. Using Xavier NX, we evaluate various model architectures to examine computational overhead differences across baseline models. Figure~\ref{fig:latency-model} confirms that \name{} achieves latency efficient performance even with TEE, primarily due to its simple weight shuffling operations specifically dedicated for attack mitigation.


\begin{figure}[t!]
    \centering
    \includegraphics[width=0.9\linewidth]{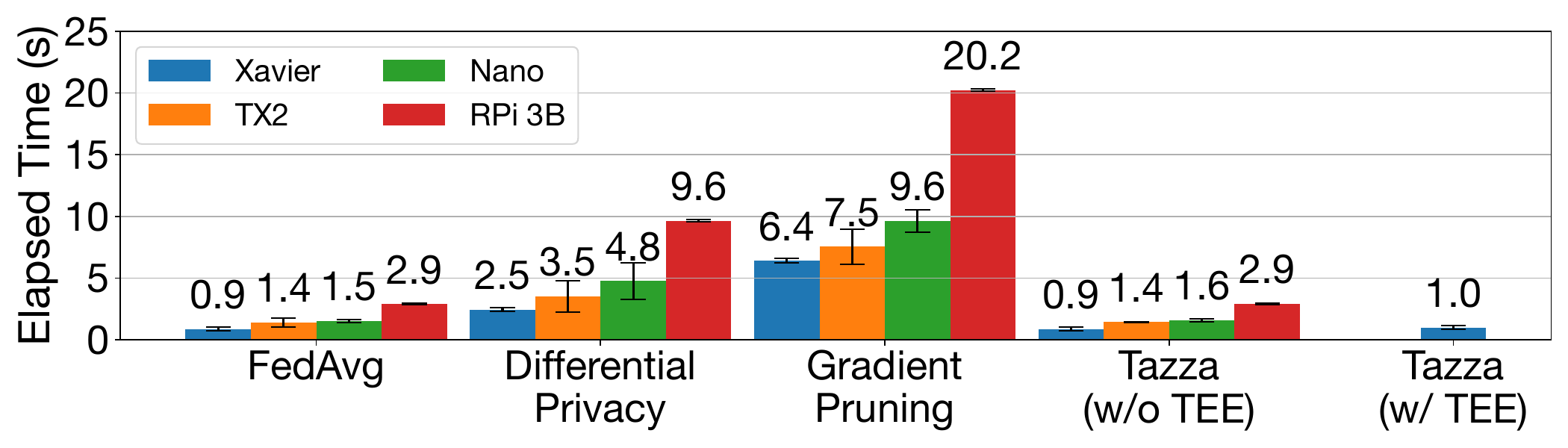}
    \vspace{-3ex}
    \caption{Latency of varying confidentiality attack mitigation schemes for different embedded platforms.}
    \vspace{-1ex}
    \label{fig:latency-device}
\end{figure}

Finally, Figure~\ref{fig:latency-device} presents the latency of various defense schemes on different platforms using a single-hidden-layer MLP model and 100 mini-batch updates. Overall, \system{} shows superior performance compared to alternatives, achieving up to 6.40$\times$ lower latency compared to gradient pruning on the Xavier NX with TEE. Note that this gain can increase up to 7.1$\times$ without TEE. Since this overhead is the price paid to mitigate attacks that aim to infer the shuffling rule via spy clients, in deployments where all clients can be trusted, our results suggest that \name{} can achieve even greater efficiency.

\subsection{Impact of Adaptive Attacks}

Lastly, we evaluate \name{} under adaptive attacks, where system-related knowledge is used to bypass defenses.

\subsubsection{Parameter-based shuffling rule inference}
\label{subsubsec:param-attack}
\begin{figure}[t]
    \centering
    \includegraphics[width=0.9\linewidth]{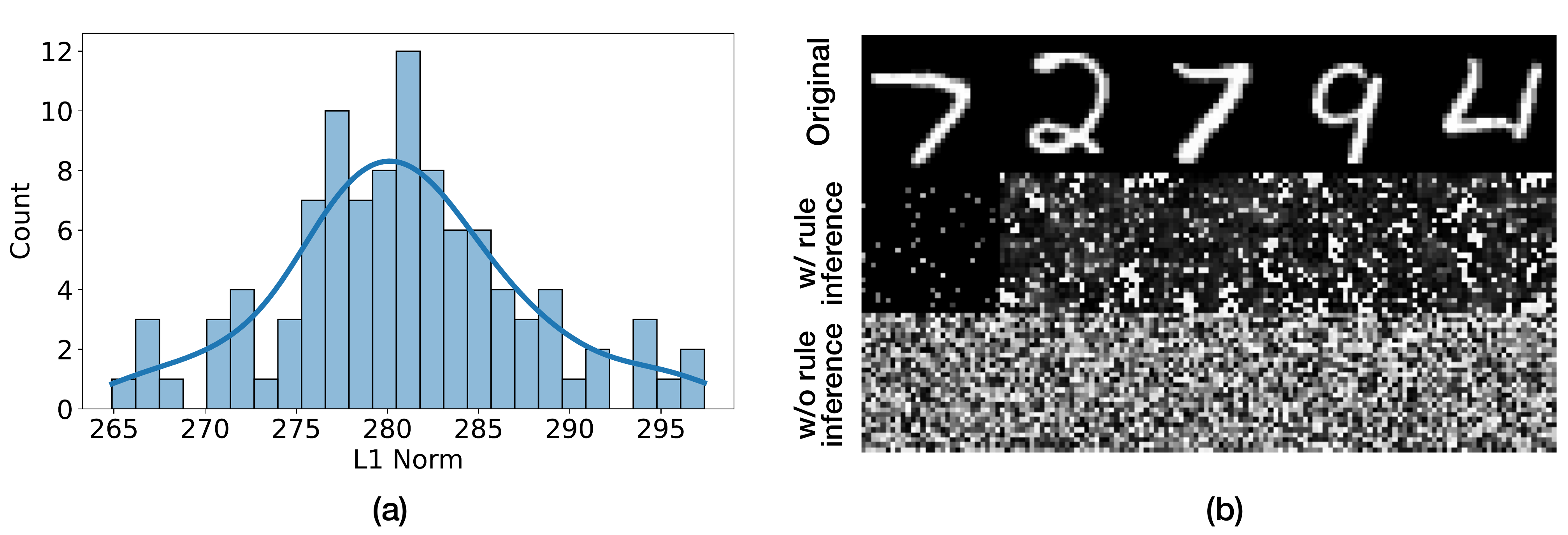}
    \vspace{-3ex}
    \caption{(a) L1-norm of server-inferred and original shuffling rule. (b) Confidentiality attack results of bottom-5 LPIPS samples w/ and w/o rule interference.}
    \vspace{-1ex}
    \label{fig:adaptive-attack-1}
\end{figure}

While \name{} targets to maintain shuffling rule confidentiality, it is practical to assume that the server may attempt counterattacks to infer this rule. For instance, the server may try to deduce shuffling rules through neural network parameter analysis. We evaluate \name{}'s robustness against such attacks using an advanced adversary employing additional shuffling rule inference techniques. Here, the attacker attempts to identify a shuffling rule that minimizes the L2-norm difference between clients' shuffled parameters and the global model.

To examine this, we used the MNIST dataset with a 3-layer MLP trained over a single update. Shuffling was applied only to the input layer, representing the simplest and most vulnerable configuration. Figure~\ref{fig:adaptive-attack-1} (a) shows the distribution of average error (L1-norm) between the inferred and ground truth shuffling rules across 100 clients. 
As the plots show, the average error is concentrated near 280, meaning that the index of a given row is off by about 280 positions. This demonstrates \name{}'s robustness against shuffling rule inference attacks. Figure~\ref{fig:adaptive-attack-1} (b) visualizes the original samples with the bottom-5 LPIPS values (i.e., 5 most successful attacks) and the corresponding attack-inferred samples with and without shuffling rule inference. As shown, the inferred samples are semantically uninterpretable, further demonstrating \name{}'s robustness against adaptive attacks that attempt to deduce the shuffling rule from model parameters.

\subsubsection{Jigsaw puzzle-based shuffling rule inference}
In addition to parameter-based rule inference, researchers have explored training models to solve jigsaw puzzles in order to reconstruct original inputs from shuffled data~\cite{li2021jigsawgan}. While these methods have shown success with a relatively small number of patches (e.g., 3$\times$3), handling the larger number of patches commonly used in neural networks (e.g., 16$\times$16, 32$\times$32) or pixel-level shuffling remains largely unexplored~\cite{carlucci2019domain}.

\begin{figure}[t]
    \centering
    \includegraphics[width=0.8\linewidth]{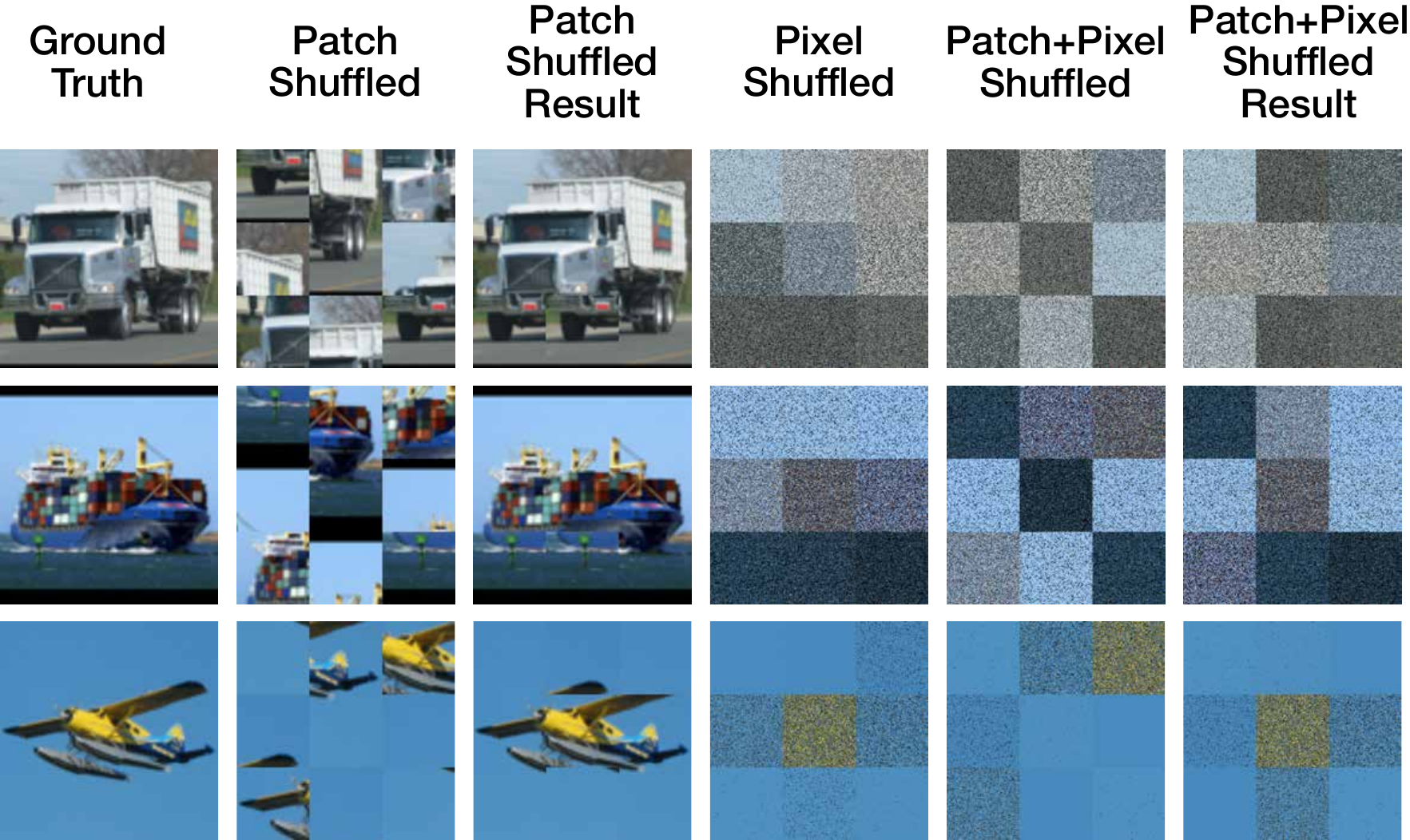}
    \vspace{-2ex}
    \caption{Results from the jigsaw puzzle solver-based image restoration attack.}
    \label{fig:adaptive-attack-jp}
    \vspace{-2ex}
\end{figure}

To evaluate \name{} on such adaptive attacks, we conduct tests using the STL dataset and a transformer-based jigsaw puzzle solver~\cite{chen2022jigsaw}. Since these models are designed for high-resolution images ($>$350$\times$350) with a small number of patches (e.g., 3$\times$3), we keep the same configuration. Figure~\ref{fig:adaptive-attack-jp} shows the image restoration attack results. The attack model partially reconstructs original images (`Patch Shuffled Result'), but when both patch- and pixel-level shuffling are applied simultaneously, the model fails to recover correct patch locations (`Patch+Pixel Shuffled Result'). Intra-patch pixel shuffling renders the semantic information unrecognizable, demonstrating \name{}'s resilience against such attacks.

Furthermore, even for simple configurations (e.g., MNIST), the number of possible shuffling rules grows factorially (e.g., (28$\times$28)!$>$$10^{1930}$), making it impractical to recover the rule via a brute-force or puzzle-solving approach. For real-world high-dimensional inputs, the shuffling rule search space expands exponentially, significantly increasing computational burden and rendering a successful attack infeasible.

\section{Discussion}
\label{sec:discussion}

\noindent \textbf{$\bullet$ Permutation Variant Neural Networks.} This work proposes \name{}, a framework addressing two core security challenges in federated learning by leveraging the permutation equivariance and invariance properties inherent in a wide range of modern neural network architectures. 
While some network structures, such as certain CNN variants, lack these properties in their native form~\cite{battaglia2018relational}, they can be effectively adapted. Specifically, CNNs can be transformed into General Matrix Multiplications (GEMM) using the standard \textit{im2col} (image-to-column) technique~\cite{chellapilla2006high, vasudevan2017parallel}. This transformation maps the spatial structure of convolutions into a permutation-equivariant matrix format, thereby enabling \name{} to secure even non-native architectures without requiring structural changes. Such optimizations highlight the versatility of designing specialized shuffling strategies; allowing a diverse set of architectures to benefit from \name{}'s mechanisms.

\vspace{1ex}
\noindent \textbf{$\bullet$ Adaptive Attacks Against \name{}.} While we evaluate two representative adaptive attack scenarios, additional strategies remain possible. For example, malicious clients could tamper with shuffling rules before uploading their models or attempt to leak rules to the server. Fortunately, \name{} is designed to identify and cluster tampered updates, and its use of shuffling rules within a TEE limits the risk of rule leakage. In rare cases where partial leakage occurs, clients can incorporate additional protections such as secure multi-party computation during rule distribution, ensuring that no single client has full knowledge of the rule without cooperation from others. While effective, these additional mechanisms introduce computational overhead. Developing lightweight yet robust security protocols to guard against adaptive threats remains a promising direction for future research.

\vspace{1ex}
\noindent \textbf{$\bullet$ Network Scalability.} Given that \name{} leverages secure peer-to-peer communication for rule exchange, scaling to massive deployments (e.g., millions of clients) poses potential communication bottlenecks. While we currently mitigate this overhead by exchanging compact random seeds instead of full matrices, ensuring robust consensus at such a scale requires further structural optimization. To address this potential bottleneck, we envision adopting \textit{Hierarchical Clustering} or \textit{Zone-based P2P} topologies as a promising technology for integration. These strategies would partition the massive client population into manageable, localized sub-groups (e.g., based on network proximity or geolocation), effectively bounding the P2P broadcast domain and ensuring scalability without compromising system robustness~\cite{ouyang2021clusterfl,liu2020client}.

\section{Conclusion}


In this paper, we introduce \name{}, a secure, privacy-preserving, and efficient federated learning framework designed to address the intertwined challenges of integrity and confidentiality in decentralized systems. Leveraging the permutation equivariance and invariance properties inherent in modern neural networks, \name{} employs innovative components such as \textit{Weight Shuffling} and \textit{Shuffled Model Validation} to enhance resilience against a wide range of attacks, including model poisoning and data leakage. Comprehensive evaluations across diverse datasets and embedded computing platforms demonstrate that \name{} effectively mitigates both integrity and confidentiality threats while maintaining high model accuracy and achieving up to 6.4$\times$ computational efficiency compared to alternative defense schemes. These results suggest the potential of \name{} as a practical and robust solution for secure federated learning.
\bibliographystyle{plain}
\bibliography{reference, eis-lab} 
\end{document}